# AI Compute Architecture and Evolution Trends

Bor-Sung Liang, *Senior Member, IEEE*

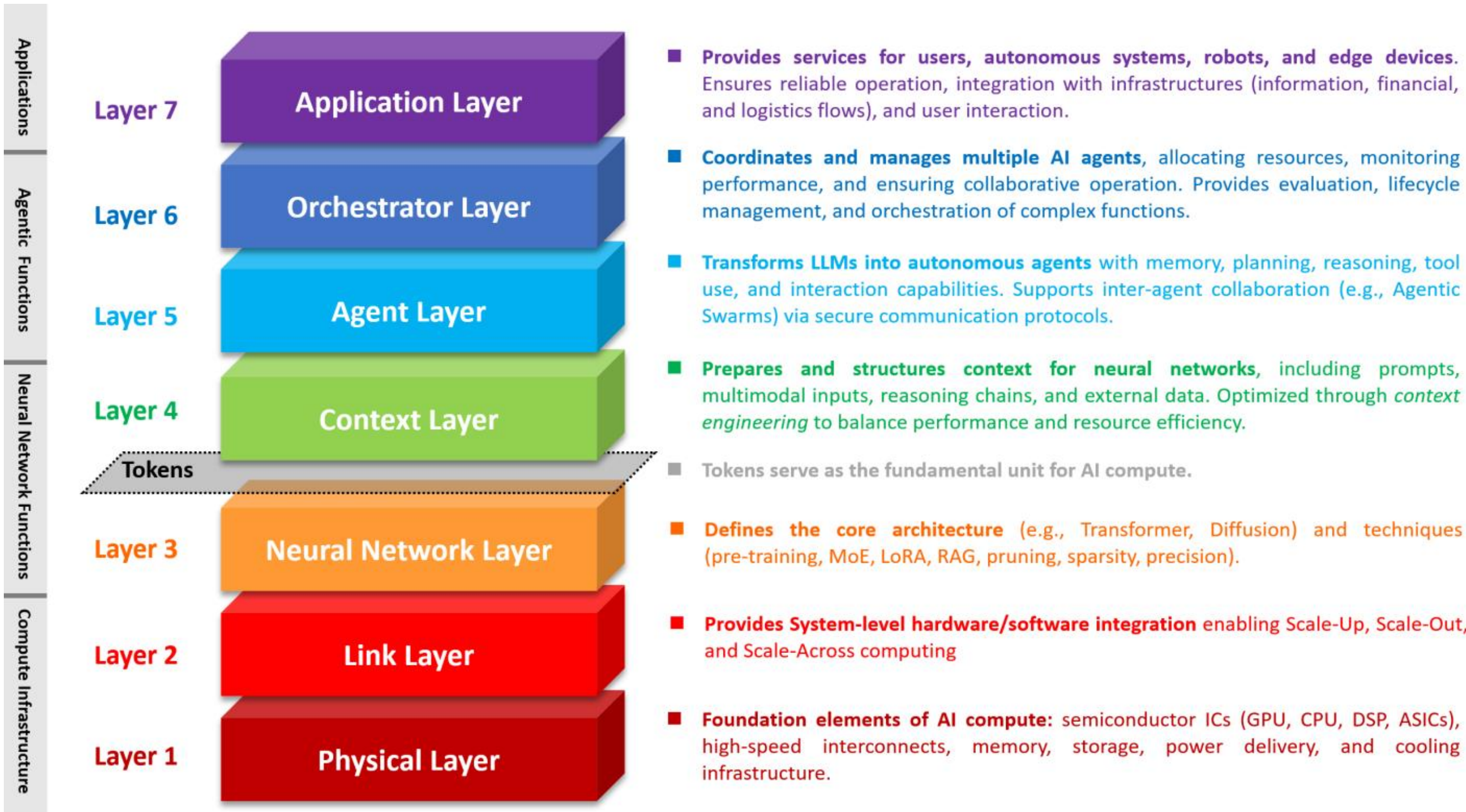

**Fig. 1** Seven-layer Model for AI Compute Architecture

***Abstract*—The focus of AI development has shifted from academic research to practical applications. However, AI development faces numerous challenges at various levels. This article will attempt to analyze the opportunities and challenges of AI from several different perspectives using a structured approach. This article proposes a seven-layer model for AI compute architecture, including Physical Layer, Link Layer, Neural Network Layer, Context Layer, Agent Layer, Orchestrator Layer, and Application Layer, from bottom to top.**

**It also explains the three stages in the evolution of large language models (LLMs) using the proposed 7-layer model. For each layer, we describe the development trajectory and key technologies. In Layers 1 and 2 we discuss AI computing issues and the impact of Scale-Up and Scale-Out strategies on computing architecture. In Layer 3 we explore two different development paths for LLMs. In Layer 4 we discuss the impact of contextual memory on LLMs and compares it to traditional processor memory. In Layers 5 to 7 we discuss the trends of AI agents and explore the issues in evolution from a single AI agent to an AI-based ecosystem, and their impact on the AI industry.**

Bor-Sung Liang is with MediaTek Inc., Hsinchu Science Park, Taiwan (e-mail: bs.liang@mediatek.com or bsliang@gmail.com). He is also concurrently serving as a Visiting Professor at CSIE (Department of Computer Science and Information Engineering) and GIEE (Graduate Institute of Electronics Engineering), EECS (College of Electrical Engineering and Computer Science) and GSAT (Graduate School of Advanced Technology) in National Taiwan University, as well as a Visiting Professor at ECE (College of Electrical and Computer Engineering) in National Yang Ming Chiao Tung University.

## I. INTRODUCTION

THE focus of artificial intelligence (AI) development has shifted from academic research to practical applications. This wave of AI advancement began with the AlexNet project in 2012 [1], which demonstrated the transformative potential of deep neural networks. Following the introduction of the Transformer architecture in 2017 [2] and the discovery of scaling laws [3], the number of model parameters and computational requirements increased dramatically, igniting a global race to develop large language models (LLMs). By 2022, the release of ChatGPT drew worldwide attention and accelerated the emergence of generative AI. More recently, AI computing has expanded further into the domains of Agentic AI and Physical AI, signaling a broader evolution toward intelligent and autonomous systems.

The rapid progress of AI holds the potential to significantly boost global productivity. If a sustainable AI-based ecosystem can be successfully established, its economic and societal impact may rival that of previous industrial revolutions. However, AI development also faces major challenges—ranging from scaling computational power and improving energy efficiency, to enhancing training and inference for increasingly complex neural networks, building effective AI agents, and establishing viable business models that sustain continuous innovation. This article provides a structured analysis of these opportunities and challenges, illustrating how AI computing has evolved into a multi-layered architecture that integrates hardware, algorithms, and intelligent systems into a coherent technological framework.

To provide a systematic understanding of how AI systems evolve and interact across different levels of technology, this article proposes a seven-layer AI computing architecture model, inspired by the conceptual clarity of the *Open Systems Interconnection (OSI) reference model* [4]. Just as the OSI model standardized communication across heterogeneous networks, the proposed framework aims to structure the complex and rapidly expanding ecosystem of AI computing.

Several prior studies have attempted to describe AI or large language model (LLM) infrastructures in layered forms [5], [6]. However, most of these works concentrate on methodologies for software systems and application development. In contrast, this work extends the abstraction to encompass the hardware, software, and system architecture dimensions, providing an integrated perspective that connects physical compute infrastructure, neural architectures, contextual intelligence, agentic behaviors, orchestration mechanisms, and application ecosystems.

The proposed seven-layer AI compute model serves as a conceptual framework for analyzing the evolution of AI computing systems, rather than a prescriptive implementation architecture. Unlike the OSI model, the layer boundaries in the proposed 7-layer model are blurry and not strictly defined, and are intended as a conceptual abstraction for analyzing AI compute evolution rather than a protocol stack. Its purpose is to organize complex cross-layer interactions—from hardware to application ecosystems—into a coherent structure that facilitates comparative analysis and future research discussions.

The proposed model consists of seven layers, arranged from bottom to top:

1. **Physical Layer** – the foundational computing elements, including hardware components, energy sources, and local interconnects.

2. **Link Layer** – the data and communication pathways that interconnect computing elements.

3. **Neural Network Layer** – the core learning and inference mechanisms of deep neural network models.

4. **Context Layer** – the representation and management of semantic and temporal context in computation.

5. **Agent Layer** – the emergence of autonomous, task-oriented AI agents.

6. **Orchestrator Layer** – the coordination and interaction among multiple AI agents and services.

7. **Application Layer** – the integration of AI capabilities into user-facing products, services, and ecosystems.

Through this seven-layer framework, we can trace how AI computing has advanced from the early era of deep learning applications to today's multi-agent and orchestrated AI ecosystems. This structured perspective also enables a more systematic discussion of the opportunities, constraints, and future trajectories in AI computing architectures.

To clarify the scope of this article, we emphasize that large language models (LLMs) are used here as a **representative use case** to illustrate the broader **evolution of AI compute architecture**. AI spans diverse domains, including computer vision, reinforcement learning, generative models, and robotics, but LLMs provide the most data-rich and visible example of how compute requirements have rapidly scaled in recent years. By tracing the evolution of LLM compute, we highlight architectural trends such as scaling strategies, numerical representation, and multi-layer integration that extend across AI domains. This framing underscores the central theme of **evolution** in our title: from accelerating a single model, toward multi-agent systems and embodied AI that integrate into the wider AI-based ecosystem.

## II. SEVEN-LAYER MODEL FOR AI COMPUTING ARCHITECTURE

Artificial intelligence (AI) systems span multiple layers of technology, from physical hardware to user-facing applications. To describe this structure in a coherent way, we introduce a seven-layer model for AI computing architecture (Fig. 1). This framework highlights how different layers of computing interact to deliver AI capabilities and provides a systematic approach to analyzing the evolution of AI technologies.

This layered perspective serves two purposes. First, it enables a structured analysis of the development trajectory of each layer, making it possible to trace technological progress from semiconductor design to AI-driven ecosystems. Second, it provides a unified framework for comparing strategies, such as *Scale-Up* versus *Scale-Out* [7], and for identifying cross-layer challenges in scalability, energy efficiency, and system integration. By adopting this abstraction, we can better interpret how AI computing has advanced and anticipate how future innovations will reshape the overall ecosystem.

**Layer 1: Physical Layer**

The Physical Layer forms the foundation of AI computing architecture. It encompasses the semiconductor integrated circuits (ICs) that execute computation, including GPUs, CPUs, DSPs, FPGAs, AI accelerators, and application-specific integrated circuits (ASICs), as well as supporting infrastructure such as high-speed interconnects, high-bandwidth memory, data storage, power delivery, and thermal management systems.

Advances in semiconductor process technology and **domain-specific architectures (DSAs)** [8] have been central to sustaining AI performance growth. While Moore's Law and general-purpose CPUs once drove progress in computing, the scale of modern AI workloads now requires specialized accelerators optimized for tensor operations and massive parallelism. These advances, coupled with innovations in memory hierarchies and advanced packaging technologies, form the critical basis for AI scalability. Nevertheless, *energy efficiency* remains a fundamental challenge at this layer, as power delivery and cooling demands have increased significantly with model size and system complexity.

**Layer 2: Link Layer**

The Link Layer enables large-scale integration of computing resources by connecting and managing thousands to millions of processing units. It includes both hardware interconnects, such as high-speed networks, switches, and optical links, and the system software stack responsible for communication, synchronization, and distributed workload management.

There are two key strategies to scale computing: **Scale-Up**, which seeks performance gains within a single compute node by (i) improving individual chips through advanced design and process technologies, (ii) integrating multiple dies within a chip package, or (iii) interconnecting multiple chips inside a compute node; and **Scale-Out**, which aggregates massive numbers of compute nodes into large-scale clusters to meet the exponential growth of AI workloads. The Scale-Out approach has become indispensable for training and deploying large language models, but it also amplifies concerns regarding latency, energy consumption, and overall system efficiency. Efficient orchestration of compute nodes, memory, and interconnect bandwidth at this layer directly determines the capability and sustainability of AI computing systems. In addition, a concept known as **Scale-Across** [9] has been proposed to extend the Scale-Out strategy across geographically distributed data centers, thereby integrating compute clusters over long distances.

**Layer 3: Neural Network Layer**

The Neural Network Layer represents the core of AI capability, defining the architectures, parameterization, and learning mechanisms that drive modern artificial intelligence. Early breakthroughs such as AlexNet (2012) [1] demonstrated the power of deep neural networks. This momentum accelerated with the introduction of the Transformer architecture (2017) [2] and the discovery of scaling laws [3], which revealed that model performance improves predictably with increases in parameter size, training data, and compute resources. These developments ignited the global race to build increasingly large-scale large language models (LLMs).

To enhance neural network efficiency and capability, numerous techniques have been developed, including pre-training, fine-tuning, low-rank adaptation (LoRA) [10] mixture-of-experts (MoE) models [11], and retrieval-augmented generation (RAG) [12]. In addition, methods such as sparsity [13], pruning [14], mixed-precision arithmetic [15], speculative decoding [16], and key–value (KV) caching [17] have been introduced to improve both computational efficiency and energy consumption. A key trend within this layer is the emergence of two distinct development trajectories. One trajectory continues to scale LLMs upward to push the boundaries of AI capabilities. The other trajectory focuses on distillation into smaller, domain-specific LLMs [18], which significantly reduce inference costs while enabling deployment on resource-constrained devices. These compact LLMs are increasingly viewed as essential building blocks for constructing AI agents, providing a balance between performance, efficiency, and accessibility. Together, these advances have shaped the practical deployment of LLMs at massive scale.

At the Neural Network Layer, information is represented and processed as **tokens**, which serve as the fundamental units of AI computation. Tokens bridge natural information and neural architectures, encompassing language units in text, image patches or feature maps in computer vision, and audio frames in speech processing. This reliance on tokenization and detokenization distinguishes AI computing from prior computing paradigms, making token-based processing a defining characteristic of this layer.

**Layer 4: Context Layer**

The Context Layer defines how information is provided to, and interpreted by, neural networks such as LLMs. Unlike traditional processors that rely on fixed memory hierarchies, AI systems depend on **context memory**—a dynamic window of tokens that captures relevant information for reasoning and response generation.

In practice, the context supplied to an LLM may include prompts, documents, images, audio, video, multimodal data, or sensor inputs. These diverse inputs are converted into tokens through **tokenization**, processed by the neural network, and then reconverted into human-readable outputs through **detokenization**. The length and richness of the context window strongly influence model performance: a larger context window typically improves reasoning and recall but also incurs substantially higher computational and memory costs [19].

A variety of methods have been proposed to optimize the use of context. Techniques such as prompt engineering [20], test-time compute [21], reasoning strategies including chain-of-thought (CoT) [22] and tree-of-thought (ToT) [23], as well as context engineering [24] [25], aim to maximize model output quality while reducing hallucinations and resource overhead. These techniques highlight that the Context Layer plays a far more proactive role in shaping neural network performance than memory management in traditional computing systems.
A useful way to illustrate this shift is by comparing traditional processor memory with AI context memory:

- **Traditional processors** play a passive role in computing. They are controlled by software programs and operate under predefined instruction sequences. Memory in this paradigm is largely static: it stores data and instructions, which are fetched and executed according to fixed control logic.

- **AI systems**, in contrast, treat context memory as an active workspace. Output tokens are generated according to the model's attention mechanism, and these tokens can also feed back to influence subsequent attention, effectively creating a self-modifying process. In other words, context memory actively participates in shaping the system's behavior, rather than serving merely as passive storage.

This distinction underscores a fundamental paradigm shift: whereas memory in traditional computing serves as a static repository, context memory in AI functions as a dynamic substrate for reasoning and adaptation.

**Layer 5: Agent Layer**

The Agent Layer converts a capable neural model, typically an LLM, into a **goal-oriented system** that can perceive context, plan actions, use tools, and interact with the external world. Beyond generating text, an AI agent maintains short-term memory and long-term memory [26], retrieves episodic/ semantic memory from external stores, decomposes tasks, selects tools or APIs [27] [28], and executes actions under constraints such as cost, latency, safety, and authorization.

In practical deployments, an agent stack commonly includes:

- **Memory**: short-term scratchpads, long-term vector or key–value stores, and preference/state profiles.

- **Reasoning & Planning**: task decomposition, reflection/ critique loops, self-verification.

- **Tool Use**: function call, API invocation, database queries, code execution.

- **Action**: web or device control, robot control, interaction.
- **Multimodality**: tokenize text, images, audio, video, or sensor streams into context, and de-tokenize for output

- **Safety, Policy, and Authorization**: role-based access, personally identifiable information (PII) filtering, credit limit control, guardrails, security.

As tasks grow in scope, multiple agents coordinate to deliver end-to-end capabilities. This naturally leads to the concept of **Agentic Swarms** [29] [30] [31], where teams of interoperable agents collaborate across organizational or platform boundaries. Inter-agent communication benefits from emerging **agent protocols**, which for capability discovery, authentication, message schemas, and trust/credit exchange, enabling standardized and auditable interactions among heterogeneous agents.

The Agent Layer thus enables LLMs to connect with other AI agents, forming swarms composed of many specialized agents that collectively provide complex functionality. Several protocols for this purpose are under active development, including Anthropic's Model Context Protocol (MCP) [32], Google's A2A (Agent-to-Agent) [33], OpenAI Swarm [34], and IBM's Agent Communication Protocol (ACP) [35].

**Layer 6: Orchestrator Layer**

The Orchestrator Layer coordinates the execution of tasks across multiple AI agents, resources, and services. While individual agents (Layer 5) can operate autonomously, orchestrators provide the **system-level governance** that enables agents to work together in scalable, reliable, and auditable ways.

At this layer, orchestration performs several critical roles:

- **Task Scheduling and Decomposition**: assigning subtasks to appropriate agents, ranking candidate solutions, and aggregating outputs.
- **Agent Evaluation**: benchmark [36] [37], selection, deployment, and tracking of performance records.
- **Resource Management**: allocating compute, memory, storage, and network bandwidth across agents to balance latency, throughput, and cost.
- **Monitoring and Control**: tracking execution status, collecting performance metrics, and providing feedback signals to improve agent selection and learning.
- **Fault Tolerance and Recovery**: detecting failures, retrying failed actions, and reassigning tasks dynamically to ensure robustness.
- **Policy Enforcement**: implementing guardrails such as safety constraints, access controls, audit logs, and compliance checks.

In modern deployments, orchestrators function much like **operating systems for agents**—they provide the substrate where heterogeneous AI services can coexist and collaborate. Emerging frameworks increasingly combine workflow engines, distributed schedulers, and reinforcement learning controllers to improve orchestration efficiency. For example, multi-agent orchestration has been studied in the context of cloud computing, robotics, and autonomous systems, and these methods are now being extended to AI agents at scale [38] [39] [40] [41].

Orchestration enables **agent ecosystems** to transition from ad hoc collections of tools into **structured, reliable services**. Orchestrators determine which agents should be activated, when they should be invoked, how their results should be validated, and how trade-offs between speed, cost, and accuracy should be resolved. This governance ensures that the overall AI system behaves predictably and meets human-centered requirements.

**Layer 7: Application Layer**

The Application Layer represents the stage where AI capabilities are transformed into real-world products, services, and ecosystems that directly impact users, organizations, and industries. It is the most visible layer of the seven-layer model, where advances in computing, neural networks, context management, and agents converge into applications that provide tangible value.

At this layer, **integration and trust** emerge as defining challenges. AI applications must align with user expectations for accuracy, safety, fairness, transparency, and accountability. Business considerations are equally critical: successful applications require sustainable economic models, regulatory compliance, and interoperability with existing digital infrastructure.

Another defining trend is the shift from single AI tools to ecosystems of orchestrated AI-based services. For example, an enterprise workflow may combine multiple agents for retrieval, reasoning, verification, and reporting, all coordinated by orchestrators to meet predefined service-level objectives. This convergence underscores the importance of **cross-layer design**: choices at the physical, link, network, and agent layers ultimately shape the reliability, efficiency, and adoption of AI applications, forming the foundation of AI-based ecosystems.

**Summary of Seven-Layer Model of AI Compute Architecture**

**Summary of Seven-Layer Model of AI**

The seven-layer AI computing architecture provides a structured framework for analyzing the evolution of artificial intelligence, from its physical computing foundations to large-scale applications. Each layer highlights a distinct set of challenges and opportunities—ranging from hardware scalability and interconnect efficiency, to neural architectures, contextual reasoning, agent behaviors, orchestration, and ecosystem-level integration. Together, these layers illustrate how advances across multiple domains converge to form modern AI systems.

By adopting this layered perspective, we can better interpret past developments, identify current bottlenecks, and anticipate future directions in AI computing. More importantly, the framework emphasizes that sustainable progress will require **cross-layer design and coordination**. The interplay between hardware, algorithms, orchestration mechanisms, and applications will ultimately determine not only the performance of AI systems, but also their trustworthiness, accessibility, and long-term societal impact—shaping the future of **AI-based ecosystems**.

It is important to clarify the maturity and evidentiary basis of the upper layers in the proposed seven-layer model.

Layers 1–4 (Physical, Link, Neural Network, and Context Layers) represent well-established technologies that are supported by extensive academic literature, industrial deployment, and quantitative performance metrics. These layers form the technical foundation of contemporary AI computing systems.

In contrast, Layers 5–7 (Agent, Orchestrator, and Application Layers) should be interpreted as emerging directions and prospective trends rather than fully matured architectural layers. While early implementations and prototypes of agentic systems, multi-agent orchestration frameworks, and AI-based applications already exist, their architectures, interfaces, and large-scale operational models are still evolving and lack the same level of standardization and empirical validation as the lower layers.

Accordingly, the discussion of Layers 5–7 in this article is intended to outline anticipated developments and design trajectories, highlighting open challenges and research opportunities, rather than to present finalized or universally adopted solutions. This distinction is made to set appropriate expectations for readers and to emphasize that the proposed framework serves as an analytical and conceptual model for understanding AI compute evolution, not as a prescriptive or complete implementation blueprint.

## III. The Evolution of LLMs (Top-Down View)

Fig. **2** conceptually illustrates the evolutionary trajectory of AI compute. Early development focused on scaling single models, exemplified by large language models (LLMs). Subsequent stages extend beyond single-model training and inference, toward agentic AI and physical AI that interact with the real world. LLMs are thus highlighted not as the sole definition of AI, but as a representative use case that makes the broader evolutionary dynamics of AI compute more concrete.

Building on this framework, the evolution of LLMs can be described in three phases. Phase 1 explores the scaling of single models through massive compute investment during training. Phase 2 emphasizes test-time compute for reasoning, while also enabling smaller LLMs that democratize AI applications. Phase 3 moves beyond single-model boundaries toward multi-agent systems (Agentic AI) and embodied intelligence (Physical AI), which together extend AI into both the knowledge and physical economies.

### *Phase 1 :Expand AI Capability Through Training Compute*

This period was characterized by exploring the capabilities of neural networks by scaling up the computational power available for AI training. Its impact spans Layers 1 to 3 of the compute architecture, and partially Layer 4 through prompting.

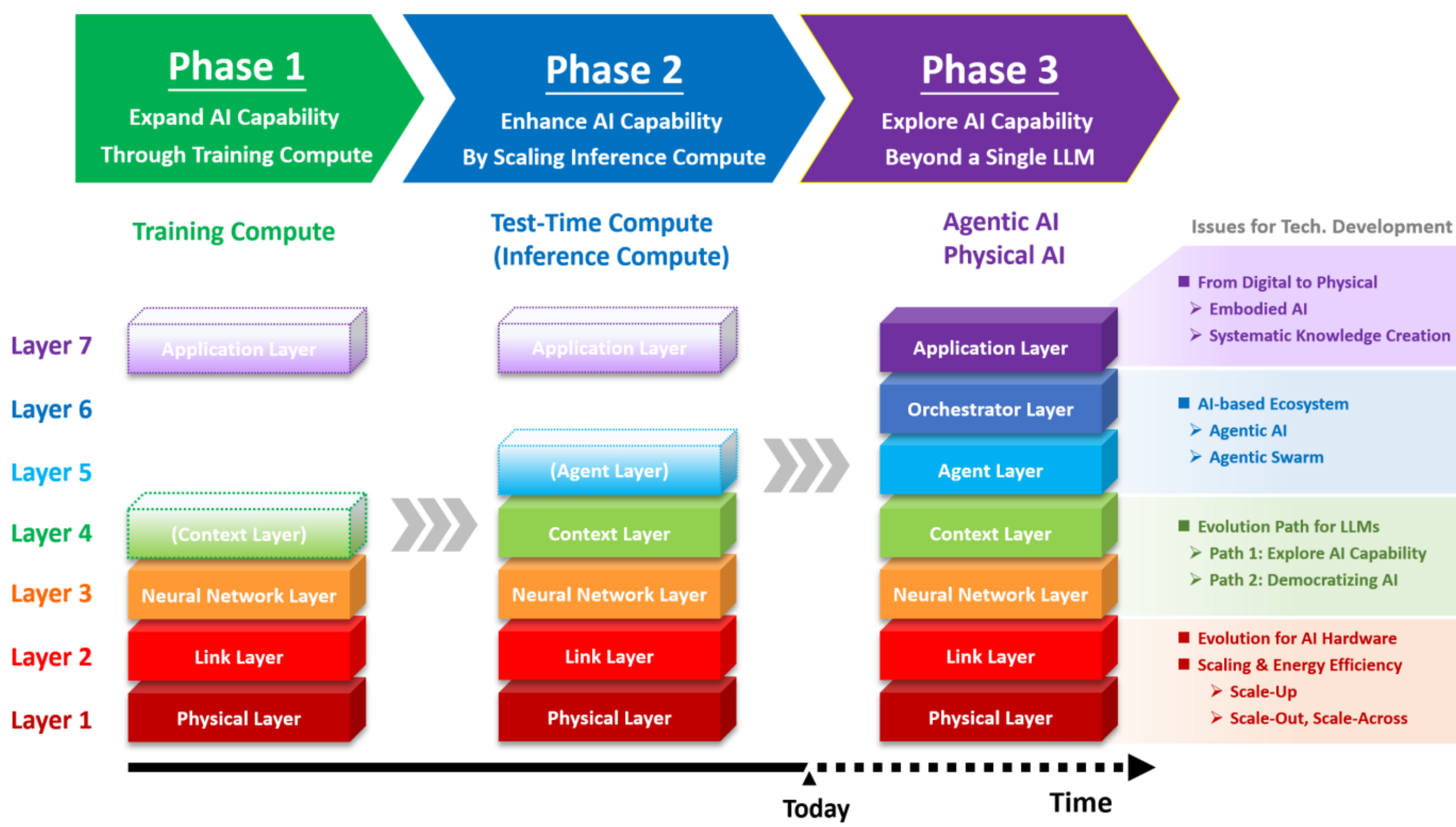

**Fig. 2** Three Phase Evolution of LLM, and Related Layers in Seven-Layer Model

Since the release of AlexNet in 2012, researchers have consistently sought to enhance AI capabilities by adding more computing power. This trend accelerated following the introduction of the Transformer architecture in 2017, alongside the discovery of scaling laws, which showed that model performance improves predictably by increasing the number of parameters, the size of training datasets, and the amount of compute. As a result, the computational requirements for training AI systems have skyrocketed. For instance, training AlexNet in 2012 required roughly $10^{18}$ FLOPs (floating-point operations), while training Gemini Ultra in 2023 required nearly $10^{26}$ FLOPs [42]. In other words, within just a decade, the compute needed for state-of-the-art models increased by nearly eight orders of magnitude ($10^8$).

Achieving this nearly 100-million-fold increase in compute within ten years could not have been accomplished through advanced semiconductor process alone. Historically, CPU performance growth driven by semiconductor scaling (Moore's Law) provided roughly a 100× increase per decade. Even with GPUs or AI-specific ASICs leveraging parallelism and tensor operations, a single chip might achieve a 1,000× to 10,000× increase per decade. Nevertheless, this still falls far short of the exponential growth required to train modern AI models. To meet these demands, the industry interconnecting chips—from dozens to hundreds of thousands, and in some cases millions—into large-scale compute clusters. This approach has driven explosive demand for advanced semiconductors.

Progress at **Layer 1 (Physical Layer)** has been fueled by advanced semiconductor processes and domain-specific architectures (DSAs). **Layer 2 (Link Layer)** plays a crucial role in enabling both Scale-Up and Scale-Out strategies, particularly by interconnecting vast numbers of GPUs and AI accelerators into distributed systems. At **Layer 3 (Neural Network Layer)**, rapid growth in neural network architectures has directly driven compute demand, with innovations such as pre-training, post-training, and fine-tuning dramatically boosting performance.

Neural networks process information in the form of **tokens**, which have become the fundamental unit of both AI computation and performance measurement. Heavy reliance on tokens distinguishes AI compute architectures from prior computing paradigms. Moreover, prompting has emerged as a critical technique for improving the performance of large-scale language models, representing an early manifestation of **Layer 4 (Context Layer)**.

*Phase 2 : Enhance AI Capability By Scaling Inference Compute*

In Phase 2, not only training compute but also **test-time compute (inference compute)** [43] emerged as a critical factor in enhancing AI capabilities. By increasing computational resources during the inference stage, neural networks can perform more sophisticated reasoning, improving their ability to handle complex tasks such as mathematics, programming, and planning. In this phase, prompting and context engineering for test-time compute became essential techniques, substantially increasing the importance of the **Layer 4 Context Layer**. In addition, LLMs began to be extended through AI agent techniques, linking this phase partially to the **Layer 5 Agent Layer** (primarily in the form of single-LLM-based agent architectures).

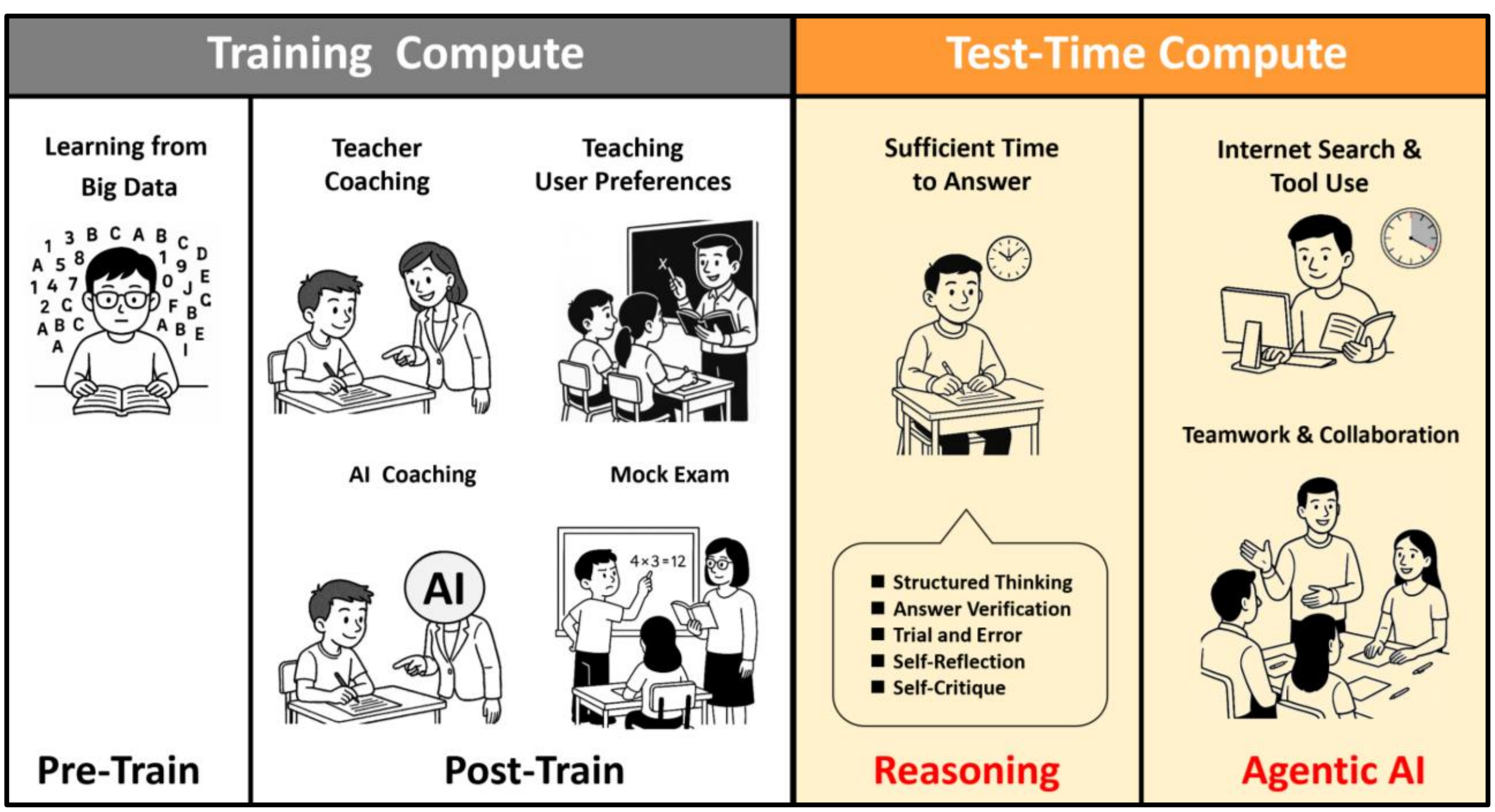

**Fig. 3** Metaphor for AI model training, including Training Compute and Test-Time Compute

During Phase 1, LLMs typically produced immediate answers after receiving a prompt. While this approach worked for simple questions, it often led to errors on complex problems, especially those involving multiple conditions or hidden dependencies. By analogy, these early models resembled students who spend significant time studying but are forced to answer exam questions instantly, without the chance to think through their reasoning. **Test-time compute** addresses this limitation by providing the model with additional computational resources and time for structured reasoning, hypothesis testing, condition verification, and answer validation. This process significantly improves accuracy for complex reasoning tasks, but it requires substantially greater computational resources. Whereas earlier improvements in AI capabilities depended almost exclusively on the training phase, they now increasingly arise from both **training** and **inference**, as illustrated in Fig. 3.

A variety of techniques have been developed to improve model performance during test-time compute. **Chain-of-Thought (CoT)** prompting instructs LLMs to reason step by step, generating intermediate reasoning traces and verifying them to arrive at more logical answers. **Tree-of-Thought (ToT)** methods extend this approach by exploring multiple reasoning paths in parallel, self-evaluating outcomes, and choosing the most promising course of action. These strategies can markedly improve results for tasks involving mathematics, logic, or planning. However, the trade-off is steep: inference compute requirements rise dramatically. Even simple text-based reasoning may consume hundreds of times more compute than earlier inference methods, while highly complex problems may require thousands or even hundreds of thousands of times more compute.

The most significant implication of this phase is the projected surge in demand for inference compute as AI usage expands. Currently, the AI ecosystem remains in its infancy, with an estimated **200–300 million daily active users (DAU)** [44] [45] [46]. Most applications are still limited to relatively simple functions such as question answering, information retrieval, summarization, programming assistance, and content generation. Truly widespread and sophisticated use of AI in work, daily life, and physical environments has yet to begin. Looking ahead, if a substantial fraction of the world's 8 billion people were to rely on AI—with billions or even trillions of agents providing services and countless robots operating across a wide range of vertical domains—the demand for inference compute would grow exponentially. As AI applications penetrate every industry and daily activity, the **frequency and intensity of inference will rise sharply**, making it foreseeable that AI inference will require extraordinarily large-scale computational resources.

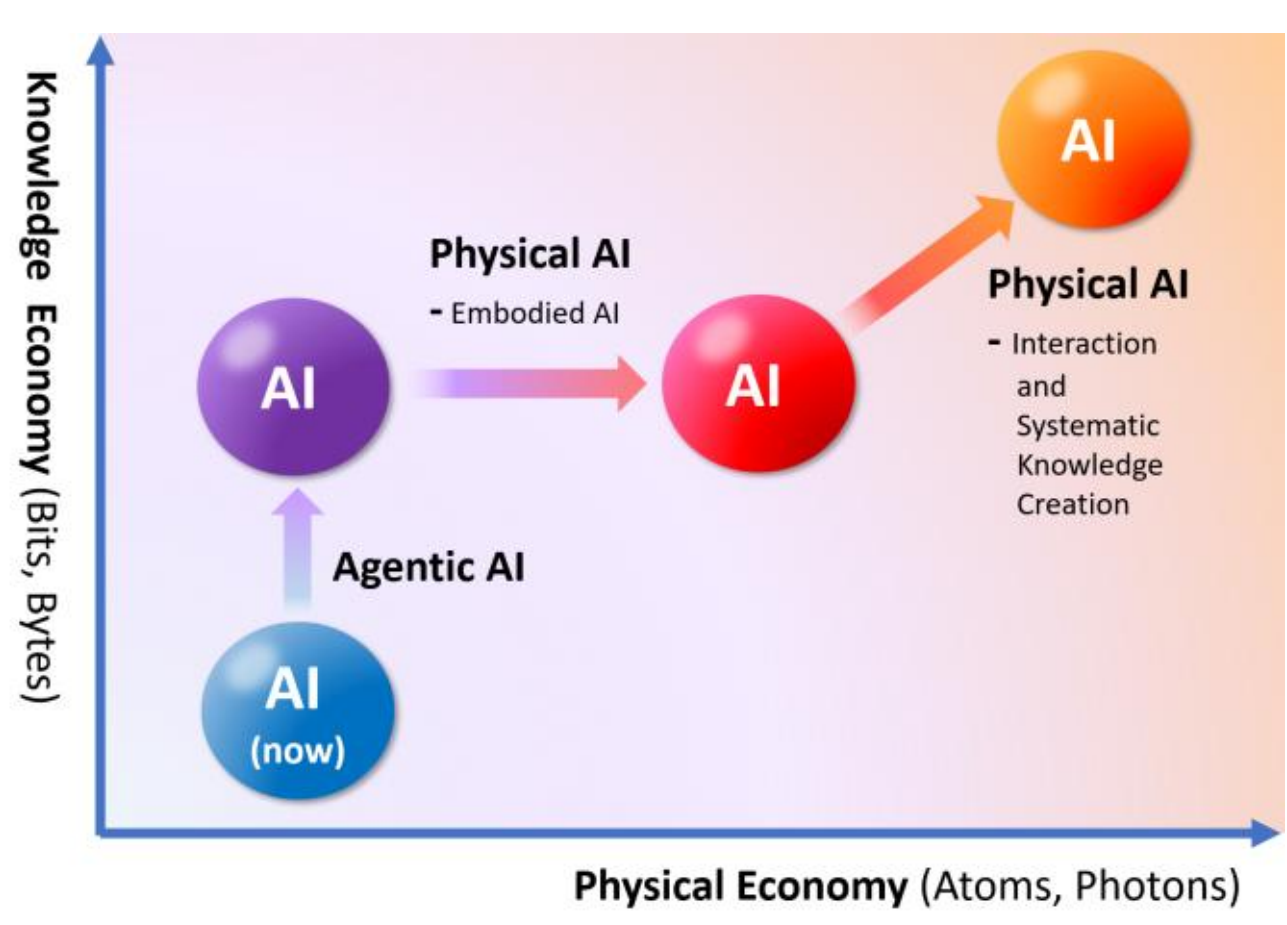

**Fig. 4** AI Evolution Trends: Agentic AI and Physical AI

*Phase 3 : Explore AI Capability Beyond a Single LLM (Agentic AI, Physical AI)*

In the previous two phases, techniques were primarily developed to enhance performance within a single LLM. By contrast, **Phase 3 aims to achieve capabilities that extend beyond the boundaries of a single model.**

There are two main trajectories. The first is to extend LLMs with additional functionalities, effectively transforming them into AI agents. This approach enhances their ability to process information, interact with external systems, and take autonomous actions—strengthening their role in the emerging knowledge-driven economy. This trajectory corresponds to **Agentic AI**, which primarily engages **Layer 5 (Agent Layer)** and **Layer 6 (Orchestrator Layer)**, where multiple agents must be coordinated at scale.

The second trajectory is **Physical AI**. Rather than focusing solely on digital knowledge processing, Physical AI expands AI's role into the physical world. It extends computing from the digital domain into the physical domain, enabling applications such as robotics, autonomous vehicles, and intelligent devices at **Layer 7 (Application Layer)**. Importantly, Physical AI not only acts upon the physical environment but also collects data through real-world interaction. This feedback loop strengthens the training process, supporting the construction of more powerful neural networks and feeding improvements back into the **Layer 3 (Neural Network Layer)**.

**Fig. 4** provides a conceptual illustration of this trajectory. The vertical axis represents contributions to the *knowledge economy* (bits, bytes), while the horizontal axis represents engagement with the *physical economy* (atoms, photons) [47]. At the lower left, today's AI (blue) remains largely confined to the digital knowledge domain. The first transition is toward **Agentic AI** (purple), where AI systems gain memory, reasoning, planning, and interactive capabilities, enabling them to act autonomously and collaborate with humans or other agents. From there, AI progresses into **Physical AI** (red), representing embodied intelligence capable of interacting directly with the physical world through robotics, sensors, and actuators. This stage bridges the knowledge and physical economies. The trajectory then advances toward a more mature form of **Physical AI** (orange), where embodied systems not only act in the world but also generate new knowledge through experimentation, observation, and continuous learning.

Together, these stages illustrate how AI is expected to evolve beyond model-centric architectures into an integrated force across both virtual and physical domains, driving future productivity and innovation. In the following sections, we will discuss Agentic AI and Physical AI in greater detail.

1) **Agentic AI**

Consider the example of a student who has graduated with excellent grades and obtained a position at a company. To ensure long-term success in their career, is it sufficient to focus solely on their academic performance? Or must we also cultivate their understanding of organizational processes, teamwork, project execution, and interpersonal communication skills?

A similar reflection applies to large language models (LLMs). When deployed in the real world, do we only need a chatbot that can provide immediate answers, or do we require an AI system capable of solving complex problems? If the goal is the latter, then it is clear that an effective AI system requires more than an LLM alone. Instead, additional capabilities must be integrated to transform the LLM into a **problem-solving system**. This is the central purpose of **Agentic AI**. In general, AI agents are LLMs augmented with complementary functions such as long-term and short-term memory, reasoning and planning, external tool use, and interaction with the environment.

A common misunderstanding is that LLMs alone represent the entirety of artificial intelligence. This overlooks the fact that users demand **applications that solve real problems**, not the models themselves. Just as people seek reliable transportation rather than an engine, users require AI systems that deliver solutions, without needing to understand the internal specifications of the underlying model. In practice, the integration of LLMs into agents will likely be handled by professional developers and organizations, who—similar to automobile manufacturers—design standardized, safe, and specialized AI agents for different purposes.

Recent work has argued that the next wave of AI may shift from monolithic frontier LLMs toward lightweight, domain-specific multimodal models and agents, motivated by both the hallucination dilemma and the rapidly growing energy cost of training and serving large models. This perspective further suggests that supporting such domain-specific agents may require rethinking hardware architectures to target extremely high energy-efficiency improvements over today's platforms [48].

In the future, therefore, professional providers will deliver a variety of agent types, each embedding LLMs with the necessary capabilities for their target use cases. This transformation marks the essence of Agentic AI.

a) Agentic Swarm

Consider the case of founding a high-tech company. Should one seek "the one"—a single, all-knowing genius who manages every aspect of the enterprise, from R&D to manufacturing, finance, marketing, and sales? Or should one instead recruit diverse specialists and assign them distinct roles to handle different functions? In practice, the latter is far more effective: organizations succeed by leveraging collaboration among experts with complementary skills, rather than relying on a single generalist.

A similar principle applies to AI systems. For most applications, there is no need to pursue a single, all-powerful "generalist AI." Instead, integrating multiple specialized AIs can yield more comprehensive and effective results.

An **Agentic Swarm** [49] consists of a collection of AI agents, each with complementary strengths, working together to achieve higher performance. This approach systematically expands AI capabilities by first converting individual LLMs into agents with enhanced functions, and then interconnecting these agents into a network-like swarm.

For example, **Fig. 5** illustrates an Agentic Swarm in the context of a travel agency. When a user wishes to plan an overseas trip, the agency's operations are managed by a swarm of AI agents specializing in customer interaction, itinerary planning, marketing, finance, supplier management, and legal compliance. While such a swarm may still collaborate with human employees for oversight and authority, it is conceivable that in the future all services could be delivered autonomously by AI agents. Planning requires coordination with external entities—airlines, hotel groups, and local service providers—each of which may themselves be managed by separate swarms of AI agents. Across the entire process, airfare, accommodation, transport, restaurants, and tickets are arranged collaboratively by agents, with purchases finalized upon user approval. Additional services, such as hiring a local guide, can also be fulfilled through specialized swarms.

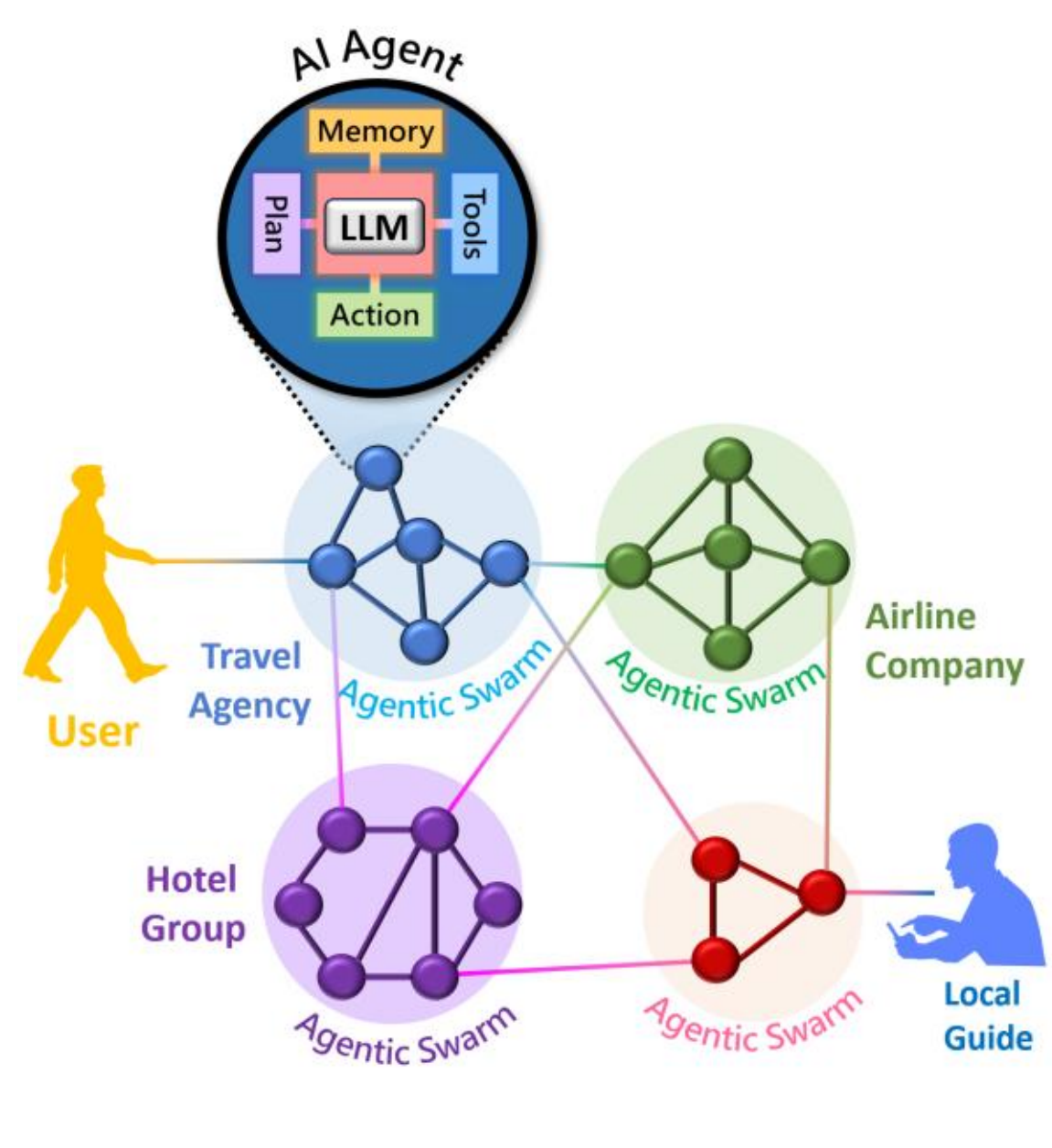

**Fig. 5** An example of Agentic Swarm

This model is highly **scalable**. Agentic Swarms can collaborate across industries—such as restaurants, railways, or retail—and they can also foster innovation through competition. Multiple swarms operating in the same sector, for example competing travel agencies, could differentiate themselves by specialization: one might focus on cultural tours rooted in fine arts and heritage, while another offers adventure tours based on expertise in mountaineering. Competitive swarms would attract more users, building reputation through demonstrated performance and reliability.

At a broader level, this architecture has the potential to **reshape entire industries**. Services and transactions may migrate into AI-driven ecosystems operated by swarms of agents, handling planning, decision-making, procurement, and commerce. Such ecosystems could eventually supplant many functions traditionally carried out through physical transactions or internet-based e-commerce.

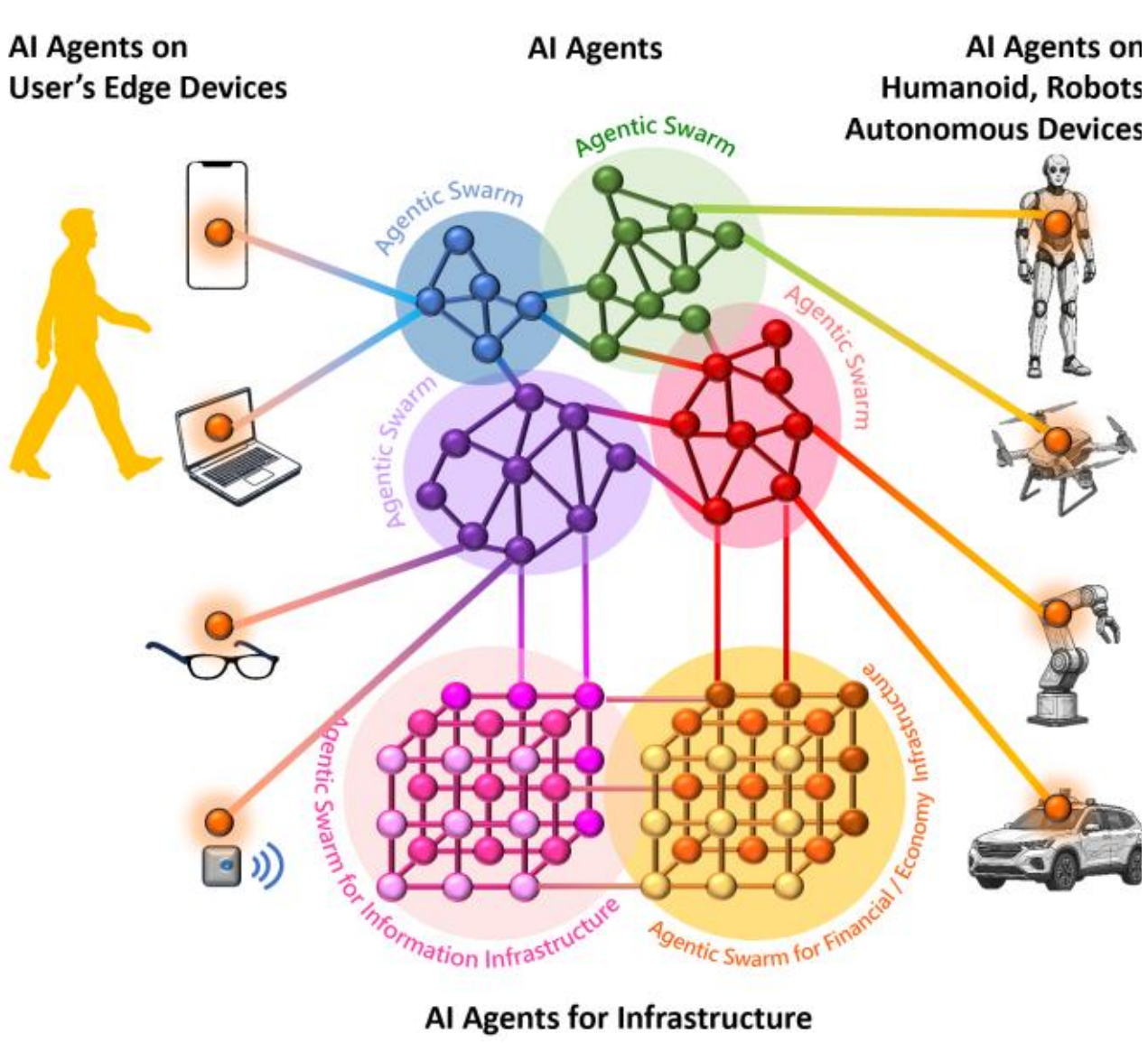

**Fig. 6** An Example of AI-based Ecosystem

b) AI-based Ecosystem

**Fig. 6** illustrates a conceptual view of the future AI-based ecosystem. In this ecosystem, diverse AI agents will be organized into different **Agentic Swarms** to provide specialized functions. It is termed an "AI-based ecosystem" because it encompasses not only AI agents but also users, their devices, robots, autonomous systems, and supporting infrastructure such as information flows, financial networks, logistics, and physical facilities. All of these elements collectively form the ecosystem.

The connections among AI agents are highly dynamic rather than permanently fixed. Links are established on demand, and decisions about maintaining persistent communication channels are made on a case-by-case basis. Furthermore, the LLMs embedded within agents may be either dedicated or shared: the same underlying LLM can be configured with different functions to create distinct agents, while multiple agents can also share a common LLM.

The operational flexibility of AI agents is a defining feature of this ecosystem. Agents and swarms may run in cloud data centers, sharing compute resources through time-sharing or parallel execution. Multiple swarms can operate concurrently within a single data center. For distributed computing, agents can be deployed on edge servers closer to users, enabling low-latency services. For security-sensitive applications, they may be deployed on on-premises infrastructure within companies, banks, factories, or research institutions. In addition, governments and enterprises will develop domain-specific infrastructures—spanning information, financial, healthcare, and public services—that form the foundation for **sovereign AI**.

AI agents will also reside on **edge devices** such as smartphones, computers, smart glasses, and other AI-enabled devices [50]. These agents act as the human–machine interface, interpreting user intent, mediating communication with external agents, and safeguarding privacy. By leveraging onboard sensors—cameras, microphones, touchscreens, accelerometers—they can assess user status, intentions, and environmental context to provide more personalized and relevant interactions. Moreover, when users initiate AI tasks, edge agents can handle **authorization and access control** before actions are executed in the broader ecosystem.

In parallel, AI agents will be embedded in **physical systems** such as humanoid robots, industrial robots, autonomous vehicles, and specialized machines. These agents will interact bidirectionally with the AI ecosystem, both receiving guidance from swarms and contributing data and actions back to the system. Collectively, such integration will blur the boundary between digital and physical domains, resulting in a pervasive AI-based ecosystem that spans virtual services and real-world operations.

2) **Physical AI**

**Physical AI** has two key implications. First, it represents the extension of AI into the physical world, often referred to as **Embodied AI** [51]. Second, it encompasses the capacity of AI systems to interact with the real world in ways that enhance their intelligence through exploration, experimentation, and feedback. Ultimately, by engaging in systematic learning and knowledge creation within the physical environment, AI can contribute to expanding human understanding of the world.

a) Embodied AI

**Embodied AI** refers to AI systems equipped with the ability to directly perceive and act in the physical world through sensors and actuators. By having a physical form, these AI agents can perceive their surroundings, take action, and collaborate with humans and other agents in real environments.

There are various types of embodied AI, including autonomous vehicles, drones, industrial robots, quadrupedal robotic dogs, and humanoid robots. Among them, humanoid robots—with their human-like appearance and kinematic structure—are particularly well suited to environments designed for humans, as they can naturally adapt to existing workspaces, tools, and workflows.

Recent surveys have highlighted the rapid evolution of embodied AI systems, emphasizing a shift from text-centric large language models toward world models that integrate perception, reasoning, and action through continuous interaction with the physical environment [52].

b) Systematic Knowledge Creation

Beyond embodiment, Physical AI also implies the capacity for **systematic knowledge creation** through direct interaction with the real world. By engaging in exploration and experimentation, AI systems can enhance their intelligence and capabilities in ways that complement data-driven training.

Current AI models are primarily trained on large datasets collected from the Internet, sensors (e.g., driving data from vehicles), or synthetic sources. They may also be trained in simulation environments, such as those developed for autonomous driving. However, under such training paradigms, the model remains effectively **isolated from the real world**. This situation resembles the thought experiment known as the **"Brain in a Jar"** [53], where the brain is cut off from direct physical interaction and can only acquire knowledge from mediated inputs. This approach fundamentally differs from learning that arises from embodied, physical interaction with the natural world (**Fig. 7**). Several limitations arise from this isolation:

- **Lack of real-world experience and interaction**: Purely data-based training (e.g., learning physical properties through videos) can help models capture basic phenomena such as motion, gravity, and Newtonian dynamics, but requires massive amounts of video data. In contrast, humans achieve such understanding far more quickly through simple physical interaction. Moreover, certain properties—such as magnetism or electricity—cannot be easily conveyed visually, limiting the effectiveness of data-only training.

- **Difficulty distinguishing fiction from reality**: Many real-world events involve intention, deception, or hidden causes. Video data alone often cannot reveal these subtleties. Furthermore, large portions of online content—such as magic shows, fantasy films, or science fiction—are explicitly fictional. Humans can leverage real-world experience to discern fact from fiction, but AI models trained solely on data struggle to make such distinctions.

- **Absence of multi-sensory and interactive input**: Human learning involves more than vision and hearing. For example, even the simple act of picking up a cup integrates tactile feedback, grip control, joint and muscle coordination, weight perception, and visual monitoring of liquid motion to avoid spillage. This multi-sensory integration is especially important for robots with high degrees of freedom (DoF), such as humanoids with 40–50 DoF, where learning from video alone is insufficient.

- **Limitations of simulation**: Simulation environments allow controlled practice and the generation of rare scenarios (e.g., hazardous driving conditions), but they remain approximations. More realistic simulations require exponentially greater compute, yet still cannot replicate the full richness of real-world phenomena. Ultimately, even the most advanced simulation is not equivalent to reality.

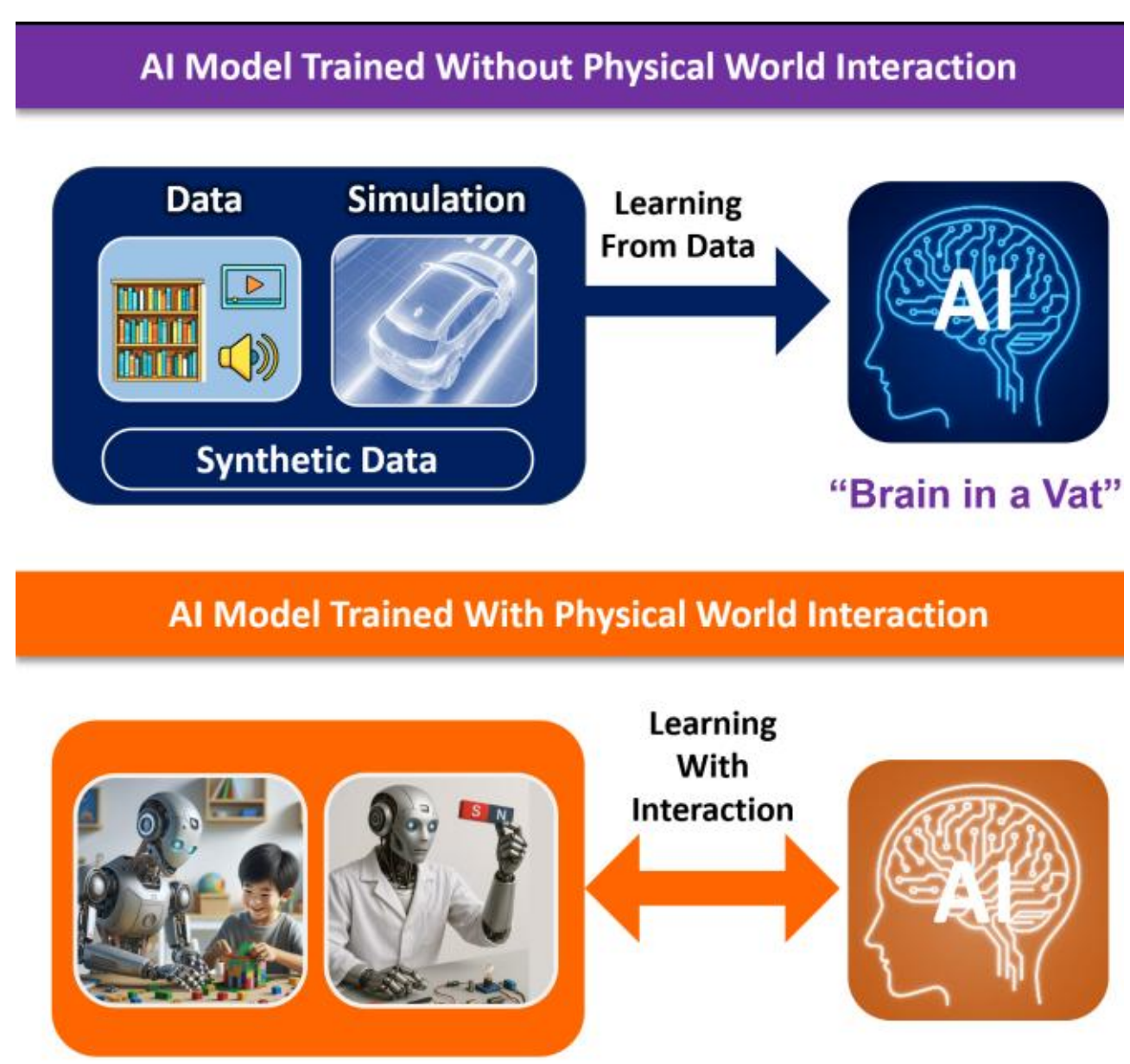

**Fig. 7** Illustration for AI Model Training with and without Physical World Interaction

Furthermore, as Kuhn emphasized in his influential book *The Structure of Scientific Revolutions* [54], **paradigm shifts** are essential to the advancement of human knowledge. Scientific understanding does not progress linearly; rather, it evolves through the discovery of anomalies that challenge existing theories, ultimately leading to new paradigms. As shown in **Fig. 8,** continuous engagement with the real world can expose such anomalies, prompting scientists to question established frameworks and thereby drive scientific progress.

For humans, the conviction that the physical world exists objectively is fundamental to this process. When anomalies arise in experiments, once we are confident that the experimental setup is sound, we tend to revise the theoretical model instead of dismissing the observed data. This openness to questioning theories rather than reality itself enables paradigm shifts and fuels scientific advancement.

By contrast, if future AI systems are trained exclusively in simulated environments without real-world interaction, anomalies observed during "experiments" would likely be attributed to imperfections in the simulation. In such cases, existing theoretical models would remain unchallenged, making paradigm shifts far less likely. Consequently, the growth of AI knowledge would be limited to the boundaries of current human science.

If, however, AI is allowed to interact with and experiment in the physical world, it could move beyond the "brain in a jar" metaphor [53]. Through embodied exploration and systematic knowledge creation, AI may help extend human understanding and accelerate scientific discovery.

**Fig. 9** illustrates the evolution of AI capabilities across Phases 1 to 3. To date, model performance has been enhanced primarily through techniques such as pre-training, fine-tuning, test-time compute, and agentic AI. Looking ahead, **Physical AI** could mark a new stage, enabling AI systems to expand beyond data-driven simulations and engage directly with the physical world. Such interaction may empower AI to assist humanity in exploring new frontiers of knowledge.

At present, however, we remain in the early stages. Agentic AI is only beginning to mature, AI-based ecosystems are still nascent, and Physical AI has been applied only in limited domains such as self-driving cars and robotics. The path toward truly interactive, knowledge-creating AI remains long, underscoring both the challenges and opportunities that lie ahead.

Phase 3 extends AI capabilities beyond individual LLMs toward system-level intelligence. Through Agentic AI, LLMs are augmented with memory, planning, tool use, and interaction abilities, forming the foundation of the **Layer 5 Agent Layer**. At larger scales, Agentic Swarms emerge, coordinated by the **Layer 6 Orchestrator Layer**, to deliver integrated services across industries. Finally, Physical AI introduces embodied and interactive intelligence, linking digital computation with the physical world and shaping applications at the **Layer 7 Application Layer**. Together, these developments represent a transition from model-centric AI to ecosystem-centric AI, pointing toward the long-term trajectory of artificial intelligence.

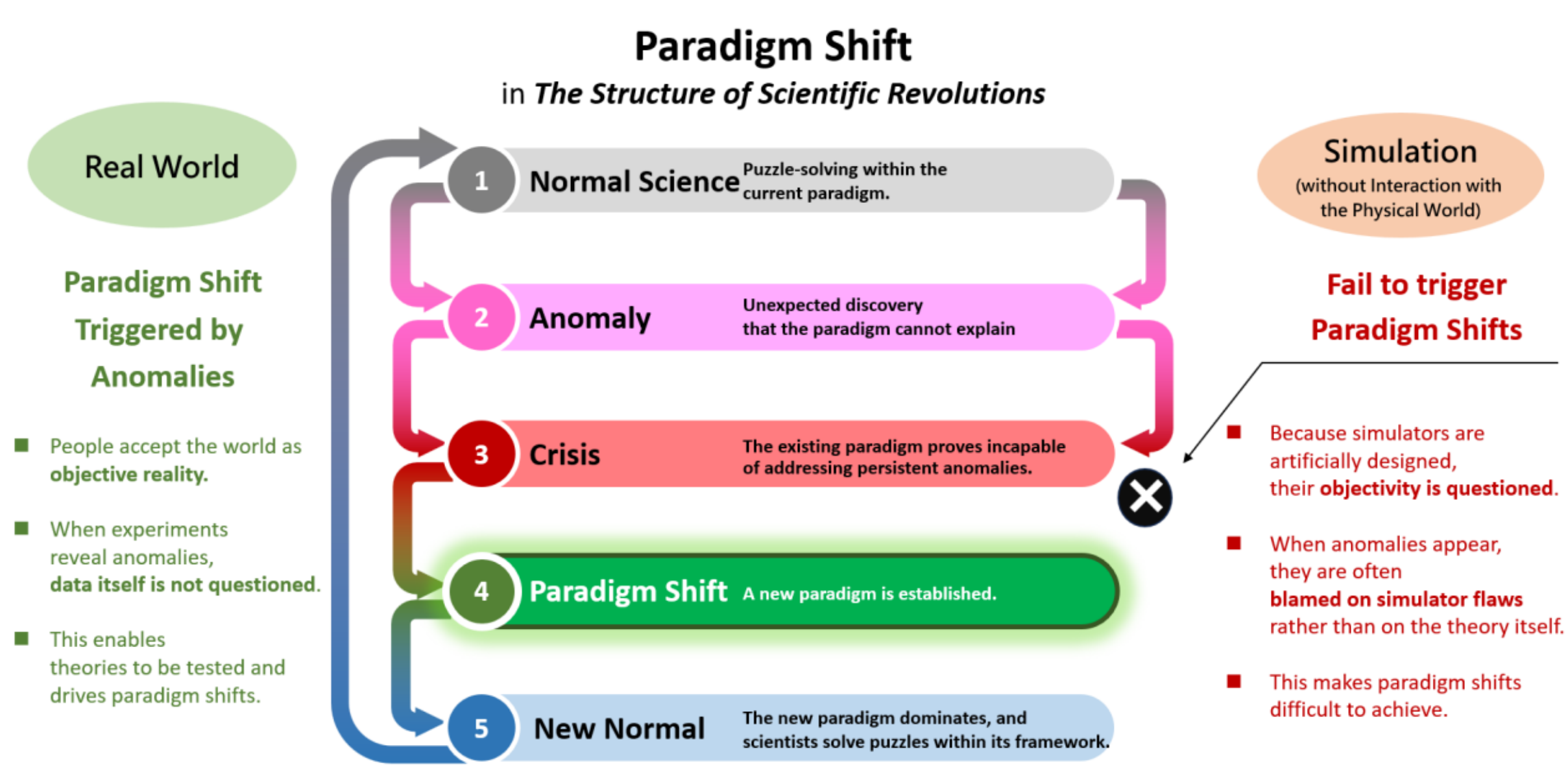

**Fig. 8** Paradigm Shifts in the Real World may be Difficult to Occur in the Simulated World.

### *D. Summary for Evolution of Large Language Model*

The evolution of large language models can be understood through three progressive phases. **Phase 1** focused on scaling training compute, driving exponential growth in model size and capability. **Phase 2** emphasized test-time compute, enabling advanced reasoning and inference through advanced reasoning and novel inference-time techniques. **Phase 3** extends AI beyond single models toward Agentic AI, swarms, ecosystems, and Physical AI, integrating intelligence across both digital and physical domains to form **AI-based ecosystems.** Together, these phases highlight the shift from **model-centric scaling** to **ecosystem-centric intelligence,** underscoring the growing importance of coordination, embodiment, and real-world interaction in shaping the future trajectory of AI.

While this section presents a top-down perspective on the co-evolution of the seven layers, the following section revisits these developments from a bottom-up viewpoint, focusing on the internal evolution and design mechanisms within individual layers. To avoid redundancy, concepts such as Scale-Up/Scale-Out, context memory, and agent coordination are introduced here only at a conceptual level to illustrate their roles in system evolution; their architectural realizations and design mechanisms are examined in detail in Section IV from a bottom-up perspective.

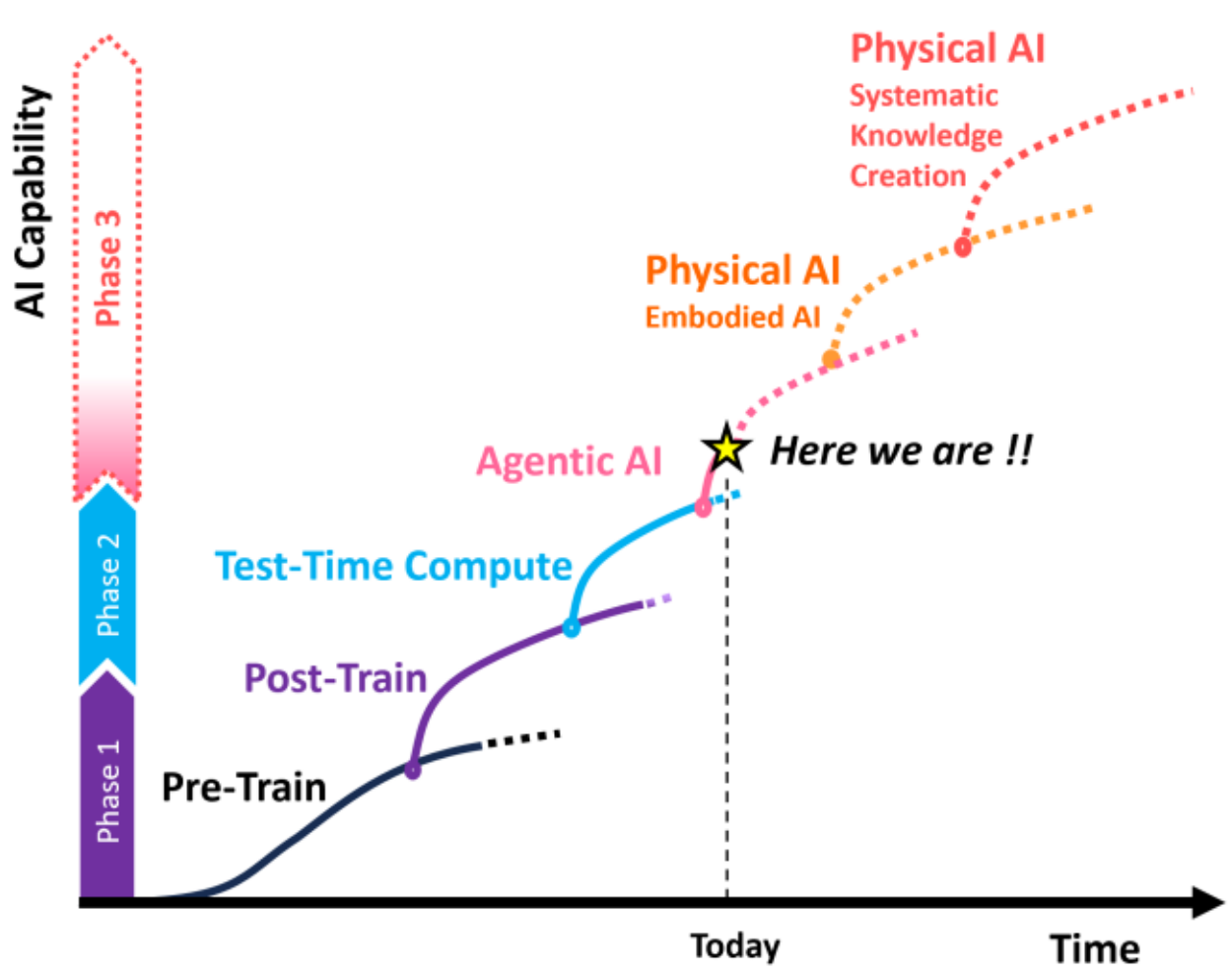

**Fig. 9** Illustration of Evolution of AI Capability (Current Status and Forecast)

## IV. EVOLUTION FOR EACH LAYERS (BOTTOM-UP VIEW)

Section III provided a **top-down view** of AI compute evolution, illustrating how the proposed seven-layer structure advanced jointly across three major phases to enable system-level capability growth. In contrast, **Section IV adopts a bottom-up perspective**, focusing on the internal evolution and design mechanisms within each individual layer, particularly Layers 1–4.

While certain concepts, such as Scale-Up/Scale-Out, context handling, and system orchestration, may appear in both sections, **they serve distinct analytical roles**: Section III emphasizes cross-layer interaction and system evolution, whereas Section IV focuses on layer-specific architectural mechanisms. Section III emphasizes how these concepts interact across layers to drive overall AI system evolution, whereas Section IV examines how similar ideas manifest as layer-specific architectural and technological innovations. Together, these complementary perspectives provide a more complete understanding of both the global trajectory and the underlying mechanisms shaping AI compute architectures.

### *A. Evolution of Layer 1 Physical Layer*

The Physical Layer provides the hardware foundation of AI computing, encompassing semiconductor integrated circuits (ICs) that execute computations. To illustrate the "Scale-Up" trajectory, **Fig. 10** shows the evolution of GPUs from 2012 to 2025, with Nvidia data-center GPUs (from K20x [55] to GB300 [56]) used as representative examples. The X-axis denotes compute throughput in TFLOP/s (tera floating-point operations per second, and 1 Tera= $10^{12}$), while the Y-axis indicates compute energy efficiency in TFLOP/s per watt. Both axes are plotted on a logarithmic scale to highlight long-term trends.

GPU evolution demonstrates continuous improvement in both computing power and energy efficiency. Four major drivers can be identified:

(1) **Advanced Semiconductor Processes.** From 2012 to 2025, GPU manufacturing nodes advanced from 28 nm to 3 nm, with 2 nm technology now available and further scaling toward the Å (Ångström) era on the horizon. These advances significantly enhanced compute performance, improved energy efficiency, reduced circuit dimensions, and increased transistor density. However, process scaling alone is insufficient to meet the rapidly growing demand for AI computing. Moreover, the pace of Moore's Law has slowed, with each successive node delivering smaller incremental gains. As a result, continued progress increasingly depends on innovations in IC design, beyond what process technology alone can provide.

(2) **Advanced Packaging.** In addition to process scaling, advanced packaging technologies have played a critical role in boosting AI computing performance [57]. High-Bandwidth Memory (HBM) has been widely adopted to increase data throughput and mitigate system bottlenecks, enabled by packaging innovations such as CoWoS (Chip-on-Wafer-on-Substrate), which provide high-density, high-speed interconnects with extremely wide data buses between compute and memory dies.

Advanced packaging also mitigates the *reticle limit*—the maximum chip area that can be fabricated in a single lithography exposure, currently around 858 mm² for standard equipment. By connecting multiple chiplets into a unified system, packaging techniques enable larger effective chip areas and greater functional integration, which is particularly valuable for AI workloads that demand massive compute resources.

Looking ahead, emerging packaging approaches such as SoIC (System-on-Integrated-Chip), SoW (System-on-Wafer), and 3DIC (three-dimensional integrated circuits) hold significant potential to further enhance compute density, bandwidth, and overall system performance. These innovations are expected to play a central role in scaling AI computing beyond the limits of traditional monolithic design.

(3) **Tensor Core Architecture.** Before the rise of AI workloads, most processor computations were executed using scalar, vector, and SIMD (single-instruction multiple-data) operations in CPUs, GPUs, and DSPs. However, AI workloads are uniquely demanding, dominated by large-scale matrix and tensor operations. This created the need for architectures that could deliver higher efficiency and throughput for dense linear algebra.

This led to the development of Tensor Core architectures, often implemented as systolic arrays [58], which perform matrix multiplications and accumulations in a highly parallel and energy-efficient manner. Tensor Cores have since become a central feature of modern AI hardware, appearing in domain-specific ASICs such as Google's TPUs (Tensor Processing Units), as well as in GPUs and other dedicated AI accelerators in ASICs. By aligning hardware more closely with the computational structure of neural networks, Tensor Cores have significantly advanced both performance and energy efficiency in AI computing.

(4) **Number Representation.** Low-bit numerical formats reduce circuit area, latency, and power consumption [59], though at the expense of numerical accuracy. Before the rise of AI computing, the dominant format was FP32 (32-bit floating point), with FP64 used in supercomputers for high-precision scientific computation. As AI workloads grew, 16-bit formats such as FP16 and BF16 (Brain Floating Point 16-bit) [60] [61] were adopted for training due to their superior speed and energy efficiency. More recently, ultra-low-precision formats such as FP8 [62] and FP4 [63] have been introduced for inference, where slight accuracy degradation is tolerable or can be mitigated through mixed-precision training. In addition, integer-only inference can be realized through quantization schemes to reduce the size to improve inference latency and throughput [64] [65]. More recently, MicroScaling (MX) techniques have been proposed to improve numerical accuracy under low-precision formats by applying fine-grained scaling to small groups of values, enabling reduced bit-width representations to better preserve dynamic range and computation fidelity [66] [67] [68]. Overall, number representation has become a critical design parameter, balancing accuracy, performance, efficiency, and implementation cost in AI hardware.

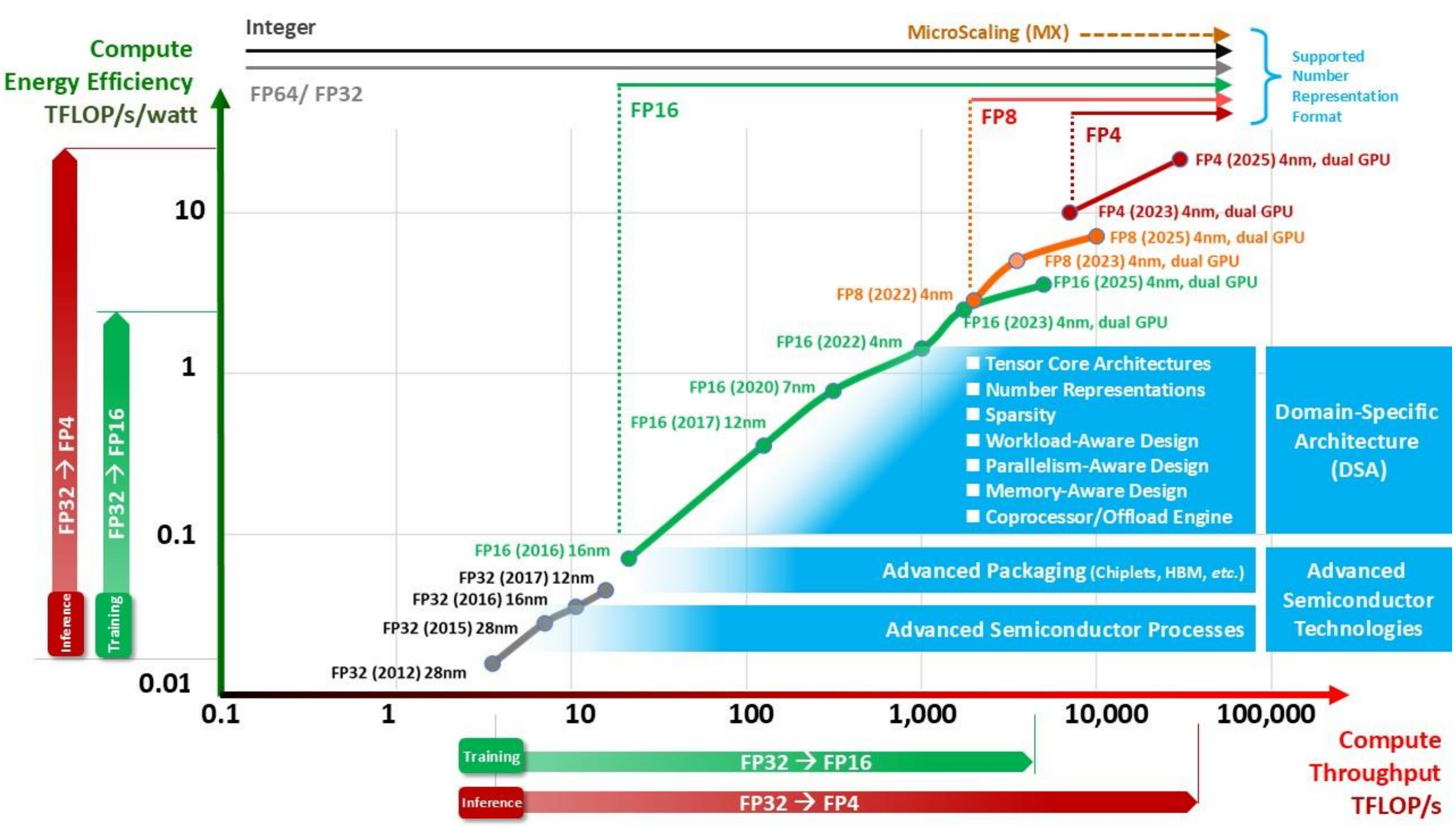

**Fig. 10** Scale-Up Evolution of Semiconductor IC (Single chip of GPU for example)

(5) **Sparsity.** Sparsity is an important property of neural networks. After training, many parameters often become zero or near zero, allowing sparsity-aware computation to significantly reduce unnecessary multiplications and memory usage. This property can be exploited to save computation and improve energy efficiency. Techniques such as *pruning* can further remove redundant connections [69]. Sparse neural networks can achieve comparable accuracy with far fewer operations [70], making them particularly attractive for efficient AI hardware.

To fully leverage sparsity for computational efficiency, hardware support is crucial. Early designs such as MIT Eyeriss [71] implemented hardware-aware sparsity acceleration, while more recent NVIDIA GPUs employ structured 2:4 sparsity (50%) patterns to double the effective throughput of dense matrix units [72] [73].. Beyond model parameters, data itself can also exhibit sparsity. For example, Google TPU's SparseCores [74] are dataflow processors that accelerate models using sparse operations to optimize irregular data workloads.

(6) **Workload-Aware Design.** Training-oriented and inference-oriented workloads exhibit fundamentally different characteristics and design strategies [75].. Training involves long-duration, high-intensity computation on pre-prepared datasets with predictable compute patterns. The primary objective is to minimize total training time, which typically leads to large batch sizes and high-throughput operation.

In contrast, inference must handle dynamic workloads where users are latency-sensitive. Consequently, inference optimization focuses on different objectives:

(i) **Interactive Serving Performance** — measuring the perceived responsiveness of the model. Common benchmarks include *Time-To-First-Token (TTFT)* and *Time Per Output Token (TPOT)* [76] [77].

(ii) **Batch Processing Performance** — monitoring system behavior under varying workloads. Benchmarks include overall throughput at different batch sizes, latency and its variance, and how these metrics scale with increasing system load [75].

(iii) **Mixture-of-Experts (MoE) Support** — MoE models employ a gating mechanism to dynamically activate only a subset of specialized experts for each input, enabling conditional computation and improved scalability. Hardware and system support enhances inference performance in MoE-based LLMs [78] by managing sparse activation patterns and distributed expert networks across devices [50].

(iv) **Deployment Support** — enabling flexible inference deployment strategies, such as *Speculative Decoding* [79] [16], which generates low-latency responses using distilled smaller LLMs and verifies outputs with full-sized models, as well as distributed inference systems spanning from data centers to edge devices [50].

(7) **Parallelism-Aware Design.** Most AI accelerators exploit parallelism at multiple levels and granularities [80].. Hardware support for parallelism is crucial to boosting performance and achieving high utilization. There are two major types: **Data parallelism** and **Model parallelism**. Common Model parallelism techniques include:

(i) **Pipeline Parallelism** — Divides a model into several stages, executing layers across multiple processing units in parallel and overlapping computation with communication to reduce pipeline bubble [81].

(ii) **Tensor Parallelism** — Distributes tensor parameters within each layer across multiple processing units, providing an additional orthogonal dimension for model partitioning [82]. In addition to direct tensor partitioning, model-level tensor decomposition techniques can further expose low-rank structures within neural network layers, reducing the dimensionality and communication overhead of tensor-parallel execution. Recent surveys have indicated that such tensorized representations are particularly effective for compressing deep neural network models, making them complementary to tensor parallelism in both large-scale and resource-constrained deployments [83].

(iii) **Expert Parallelism** — Distributes experts in *Mixture-of-Experts (MoE)* architectures across processing units to support efficient parallel execution [84].

(8) **Memory-Aware Design.** In AI computing, memory is as crucial as computation itself. If data cannot be supplied in time, compute units remain idle regardless of processing capability. Hence, memory hierarchy, capacity, bandwidth, and scheduling are key considerations in hardware architecture. Memory-aware design techniques include:

(i) **Cross-Hierarchy Memory Management.** Beyond on-die memory, on-package memories such as High Bandwidth Memory (HBM) [85] [86] and emerging Compute Express Link (CXL)-attached memory [87] [88] provide high-capacity, high-bandwidth options. Efficient management across these heterogeneous memory tiers, including off-chip memory, is essential to optimize performance.

(ii) **Cache and Context Management.** Optimization methods include pre-fill [89] and pre-fetch [90] techniques for key–value (KV) caches, cache offloading [91], and dynamic resource management for varying context lengths [92].

(iii) **Compression.** Memory compression techniques include cache compression [93] and model compression through quantization, pruning, knowledge distillation, and low-rank factorization [94]. Further compression can be applied to model weights and activations in LLMs [95].

(iv) **Compute-Near-Memory and Compute-in-Memory.** The data movement bottleneck between processors and memory — known as the "memory wall" [96] or the "von Neumann bottleneck" [97] — is particularly severe for LLM workloads with massive parameters and data transfers. Research efforts have explored Compute-Near-Memory (CNM) and Compute-in-Memory (CIM) architectures to mitigate these limitations [98] [99].

**(9) Coprocessor/ Offload Engine.** When a single AI accelerator faces excessive computational or memory demand, additional coprocessors or offload engines can complement the main chip. In large-scale inference, the context length can exceed one million tokens, causing distinct bottlenecks: the context phase is compute-bound, while the generation phase is memory-bandwidth-bound. To address this, specialized **GPU variants** such as NVIDIA Rubin CPX [100] target specific workloads. Other strategies include **CPU offloading** for activation data [101], and **CXL-based offloading** for memory extension. Because Compute Express Link (CXL) allows devices to access each other's resources without host intervention, model parameters or activations that exceed GPU memory can be offloaded to CPU-attached CXL memory, effectively mitigating memory bottlenecks [91] [102].

**(10) Analog AI Computing.** While the evolution of the Physical Layer discussed in (1)–(9) has been largely driven by digital AI computing, grounded in domain-specific architectures and advanced semiconductor technologies, research efforts have also begun to explore analog AI computing as an alternative approach for improving energy efficiency in neural network workloads. As summarized in recent tutorial and review literature [103]. By exploiting physical device properties such as voltage, current, or conductance for computation, analog architectures enable highly parallel multiply–accumulate (MAC) operations, thereby achieving significant gains in energy efficiency for selected workloads. Recent circuit- and system-level demonstrations have further shown the feasibility of robust analog compute-in-memory macros and analog acceleration of selected large language model (LLM) kernels, such as attention mechanisms [104] [105].

Despite these promising results, analog AI computing faces fundamental limitations when compared with **digital AI computing**, which remains the dominant paradigm for large-scale AI systems. Digital representations offer **high accuracy, reproducibility, and portability**, allowing models trained on one platform (e.g., GPUs or ASICs) to be easily transferred, modified, and deployed across heterogeneous inference hardware, including edge devices. In contrast, analog AI models rely on physical device states, making precise replication, post-training modification, and cross-platform deployment more challenging. Consequently, analog AI computing is best viewed as a **complementary computing regime** for ultra-low-power and highly specialized workloads, rather than a replacement for flexible, scalable, and ecosystem-driven digital AI computing.

**(11) Weightless Neural Networks.** Weightless neural networks (WNNs) represent an alternative computing paradigm that departs from conventional deep neural networks based on weighted multiply–accumulate operations. Instead of storing knowledge in continuous-valued weights, WNNs encode information using **memory-based structures**, such as lookup tables or associative memories, thereby eliminating explicit multiplications during inference.

Recent work has demonstrated that WNNs can enable energy-efficient and low-latency inference for edge and embedded platforms due to their memory-centric execution and reduced arithmetic complexity [106]. More recently, differentiable weightless neural networks have been proposed to bridge classical WNN models with modern gradient-based optimization, improving training flexibility while preserving inference-time efficiency [107].

Despite these advances, the representational capacity and scalability of WNNs remain limited compared with modern deep neural networks, particularly large language models. Consequently, WNNs are best viewed as a **complementary design point** within the Physical Layer, well suited for specialized, ultra-low-power inference tasks rather than as a replacement for scalable digital AI computing architectures.

**(12) Quantum Computing-Based AI.** Quantum computing-based AI represents a fundamentally different and long-term computing paradigm that may influence the future evolution of AI computing at the Physical Layer. By exploiting quantum-mechanical phenomena such as superposition and entanglement, quantum computing has been investigated as a potential accelerator for selected learning, optimization, and sampling problems.

Recent survey and review studies provide a comprehensive view of quantum machine learning (QML), covering its foundations, representative algorithms, software frameworks, and emerging applications, while emphasizing the constraints imposed by current noisy intermediate-scale quantum (NISQ) hardware [108] [109]. At a more fundamental level, advances in tensor network representations have indicated the structural connections between quantum computing and machine learning [110]. Despite these advances, quantum computing remains constrained by limited qubit counts, noise, and immature software ecosystems, preventing large-scale AI training or deployment comparable to modern digital AI platforms.

From an architectural perspective, quantum computing-based AI should therefore be viewed as a speculative and long-term complementary trajectory, rather than a replacement for digital AI computing. Its primary significance lies in expanding the design space of the Physical Layer as AI workloads continue to challenge the limits of classical scaling and energy efficiency.

Among these technologies, advanced process nodes and advanced packaging represent key innovations in semiconductor manufacturing, while tensor core architectures, number representations, sparsity, workload-aware, parallelism-aware, memory-aware, and coprocessor/offload engine designs fall under the domain of **domain-specific architectures (DSAs)**. DSAs enable joint hardware–software–system co-optimization for AI algorithms and have largely defined the trajectory of modern AI accelerators, distinguishing them from traditional general-purpose processor architectures. Consequently, DSAs remain the dominant and most practical approach for scaling AI performance in contemporary systems.

Recent surveys on LLM-centric computing systems have shown that most performance and efficiency gains in contemporary AI accelerators are achieved through collaborative hardware–software co-design, encompassing algorithm-level optimizations, system-level techniques, and domain-specific accelerator architectures [111]. More broadly, other survey studies have framed machine learning hardware design as a joint optimization problem across efficiency, flexibility, and scalability, highlighting architectural choices, dataflow organizations, sparsity exploitation, and system-level integration as key design dimensions [112].

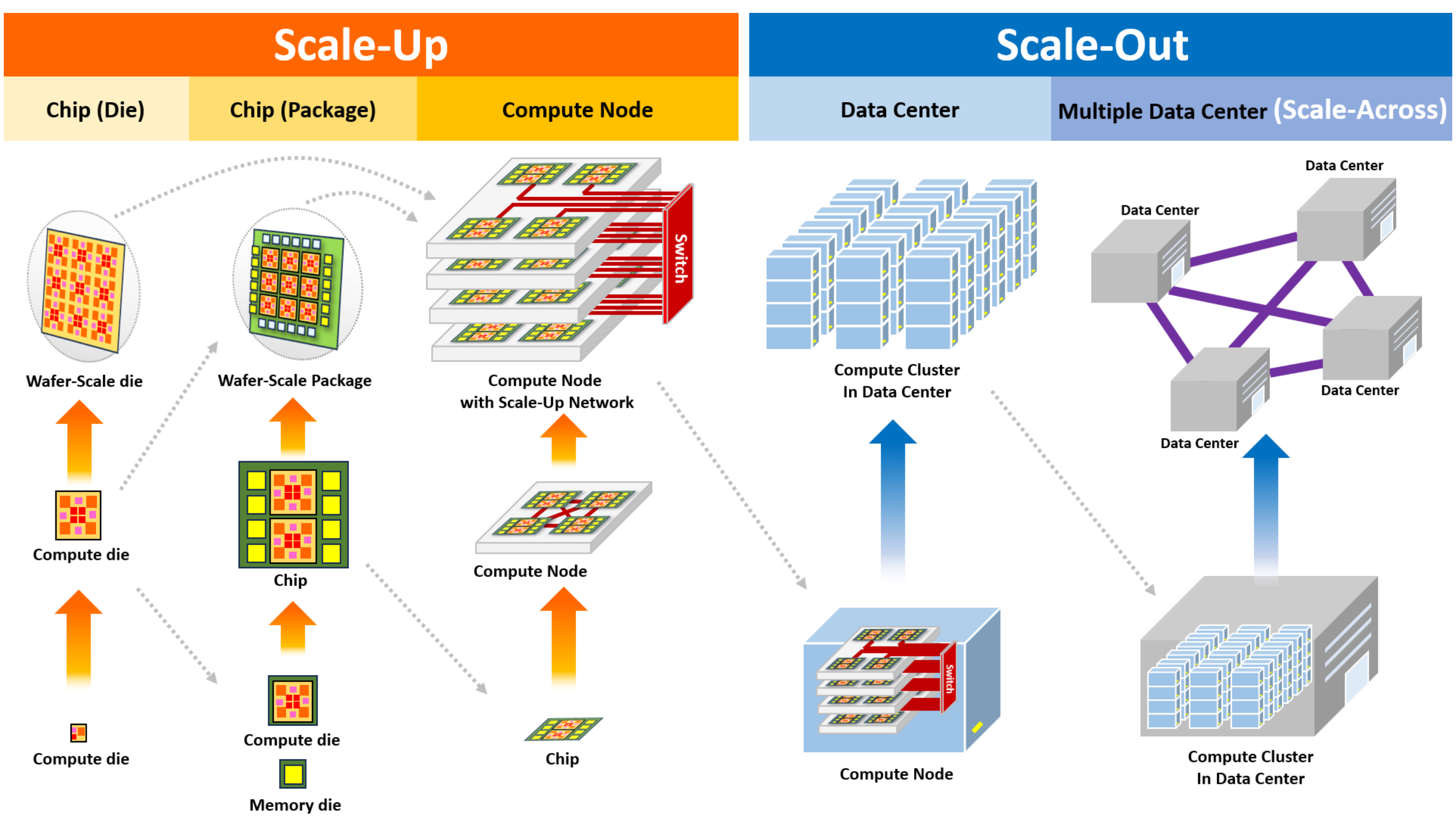

**Fig. 11** Scale-Up, Scale-Out Strategies for expanding AI computing capacity

As shown in **Fig. 10,** between 2012 and 2025, FP16 computation—widely used in AI training—delivered roughly a 1,000× increase in throughput and a 100× improvement in energy efficiency per GPU. In contrast, ultra-low-precision formats such as FP4, primarily applied in inference workloads and combined with arithmetic and architectural optimizations, have the potential to achieve more than a 10,000× gain in throughput and a 1,000× gain in energy efficiency per device, provided that accuracy remains acceptable. These results underscore how design techniques within the DSA space exert a decisive influence on overall system performance.

Beyond data-center and conventional edge accelerators, recent work has demonstrated that AI inference can be deployed on ultra-resource-constrained edge devices through aggressive co-design across neural architectures, quantization, memory scheduling, and lightweight runtimes. Such progress highlights how extreme energy-efficiency–driven Physical Layer and architectural optimizations can substantially broaden the deployment envelope of AI systems [113].

At the same time, the gap between throughput scaling and energy-efficiency scaling remains evident. Energy efficiency has improved more slowly than throughput, trailing by approximately an order of magnitude on a logarithmic scale. This growing disparity highlights a fundamental challenge for sustainable AI hardware development and helps explain the increasing research interest in alternative computing paradigms, such as analog computing, weightless neural networks, and quantum computing, discussed in (10)–(12). While these approaches are unlikely to replace DSAs in the near term, they expand the design space of the Physical Layer and offer complementary perspectives on addressing the long-term limits of energy-efficient AI computing.

*B. Evolution of Layer 2 Link Layer*

The Link Layer comprises the system hardware and software that interconnects and manages the compute elements in Layer 1. To meet the massive demands of AI workloads, both Scale-Up (vertical performance improvements) and Scale-Out (horizontal performance expansion) strategies are required. In practice, large-scale AI systems increasingly rely on Scale-Out, with modern data centers already connecting hundreds of thousands of chips—and potentially millions in the near future—to deliver the required compute capacity.

**Fig. 11** illustrates the architectural strategies of Scale-Up and Scale-Out for expanding AI computing capacity. On the left, the Scale-Up approach focuses on enhancing performance within a single compute node. This begins at the chip level, where chip dies can be scaled through DSA design and advanced semiconductor processes, and then multiple compute dies and memory dies can be integrated into larger chip packages [57] or even wafer-scale devices [114] [115]. These chip dies are then interconnected within a package through high-bandwidth interconnects, further extended by scale-up networks inside compute nodes [116]. On the right, the Scale-Out approach expands capacity by linking multiple compute nodes into large-scale clusters within a data center, and further across multiple data centers (Scale-Across). Scale-Out provides extreme performance through massive parallelism, but often at the cost of greater energy consumption and communication overhead. Together, these strategies, Scale-Up improving chip and node-level performance, and Scale-Out extending to system and cross-data-center integration, form the foundation of modern hyperscale AI infrastructure.

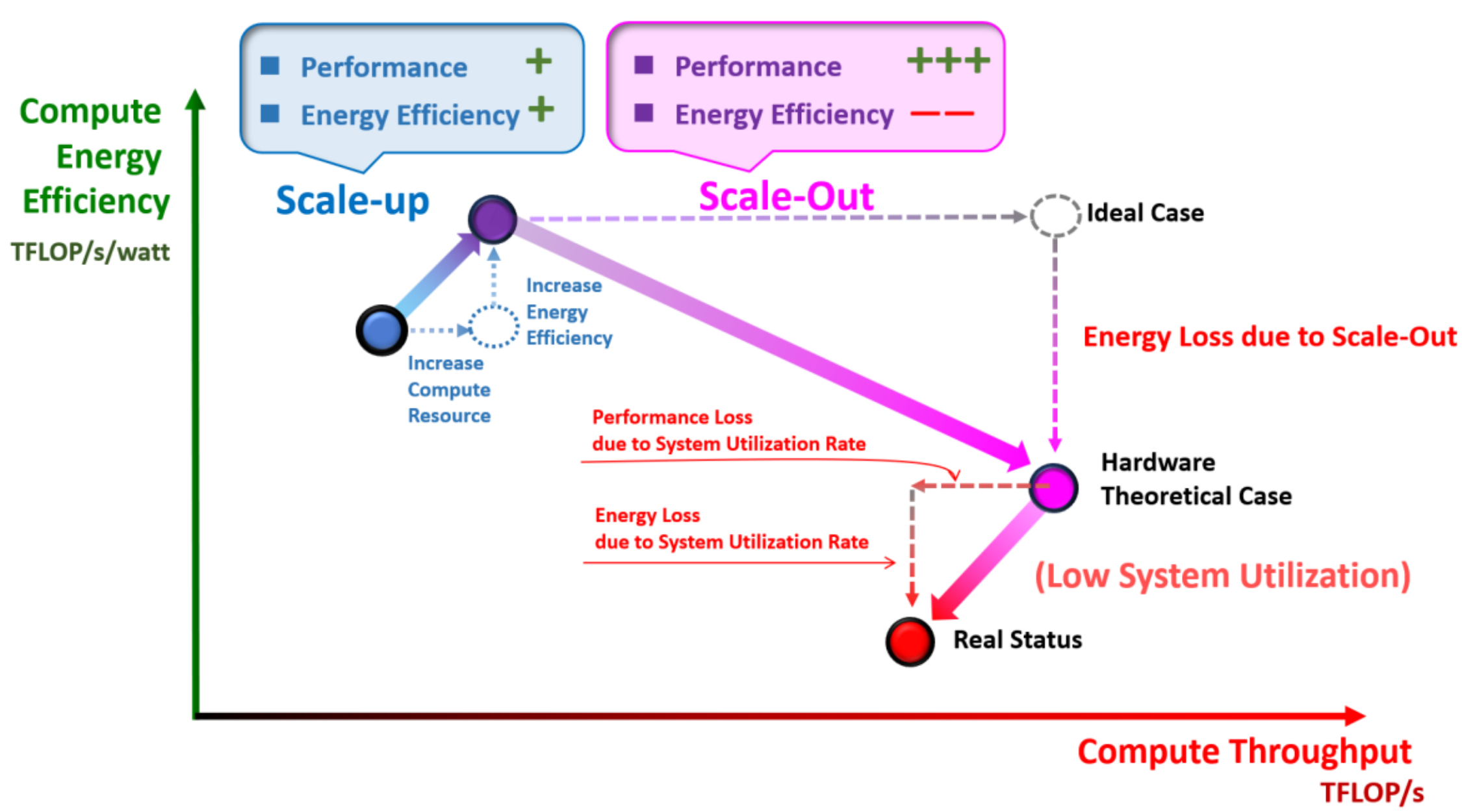

**Fig. 12** The Impact of Scale-Up and Scale-Out on AI Computing Performance and Energy Efficiency

Achieving this scale requires a combination of sophisticated architectural design, high-bandwidth interconnects, robust software frameworks, efficient power and cooling systems, and comprehensive operational strategies. Building hyperscale AI data centers is therefore an extremely complex engineering challenge.

1) **Impact of Scale-Up and Scale-Out on AI Computing Power and Energy Efficiency**

Scale-Up and Scale-Out both increase computing capacity, but they also impose additional system burdens, often reducing overall energy efficiency (**Fig. 12**). The analysis is as follows:

**(1) Scale-Up.** Scaling up enhances performance through three key approaches: (i) improving individual chips via advanced process technologies and design innovations, (ii) integrating multiple dies within a single package, and (iii) interconnecting multiple chips inside a compute node, as illustrated in the left part of **Fig. 11**. Energy efficiency may also benefit from semiconductor process advances and domain-specific architectures (DSAs) in IC design, as shown in **Fig. 10**. In addition, optimized high-speed interconnects between dies and chips are critical since they directly impact both performance and efficiency".

**(2) Scale-Out.** Scaling out expands computing capability by linking large numbers of chips—ranging from dozens to hundreds of thousands, and potentially millions in the future—across compute nodes, racks, clusters, and even geographically distributed data centers. This horizontal expansion dramatically amplifies aggregate performance but introduces challenges: high-speed interconnects for node-to-node, rack-to-rack, and cluster-to-cluster communication consume substantial energy, and increased data movement across long interconnect paths adds latency bottlenecks. Thus, scale-out delivers raw performance gains but often at the expense of energy efficiency. Technologies that mitigate the energy overhead of large-scale interconnection are essential for sustainable AI progress.

**(3) System Utilization.** Even when theoretical capacity is increased through Scale-Up and Scale-Out, real-world utilization often falls short. Data dependencies and delivery delays create "bubbles" in the compute pipeline, leading to idle cycles that waste energy and reduce throughput. To address this, techniques such as caching, batching, pipelining, and prefetching [116] [117] [118] are used. For example, data centers may aggregate tasks across workloads via batch processing to improve GPU or ASIC utilization—though at the cost of higher latency per task. Ultimately, utilization is a decisive factor in determining how much of the theoretical performance and efficiency can actually be realized at scale.

**(4) Software Frameworks.** In addition to hardware scaling, software frameworks that orchestrate multi-GPU and multi-node systems are equally critical to the effectiveness of the Link Layer. For training, frameworks like Megatron-LM [119] enabled efficient scaling of LLMs to thousands of GPUs through tensor, pipeline, and data parallelism. Building on this foundation, state-of-the-art systems now extend scaling to tens or even hundreds of thousands of GPUs, enabling the training of trillion-parameter models. For inference, inference platforms like Dynamo [120] provide a high-throughput low-latency framework for deploying AI models at scale in distributed environments. In addition, the Key-Value (KV) cache, storing past keys and values for the attention mechanism, emerges as a critical bottleneck in both memory usage and latency. Many researches toward KV-cache, such as vLLM [121], an LLM serving system for near-zero waste in KV-cache memory and flexible sharing of KV cache, reducing KV-cache size with cross-layer attention [122], layer-condensed KV-Cache [123] and KV-cache compression [124]. In addition, disaggregating the prefill and decoding stages has emerged as an effective and increasingly adopted system-level technique for large language model serving, enabling more efficient resource allocation and improved goodput under distributed inference workloads [125]. Overall, the synergy of hardware and software in Layer 2 advances the compute platform, enabling the practical deployment of large-scale AI across both training and inference stages.

As shown in **Fig. 12**, performance and energy efficiency are both critical parameters in the Scale-Up process, as they directly define system capability. While Scale-Out substantially increases aggregate performance, the power consumption of inter-node communication—whether between racks, across clusters, or spanning data centers—imposes significant limits on energy efficiency. Consequently, technologies that reduce the energy cost of large-scale interconnection will be essential for sustaining progress in AI computing.

This trade-off highlights the central role of the Link Layer: it must deliver scalable, reliable, and energy-aware interconnect architectures to support the next generation of AI systems.

**2) Evolution of AI Compute**

**Fig. 13** illustrates several categories of AI computing systems to highlight their evolution over time. Five categories are shown:

- **GPUs** (e.g., Nvidia GPUs [55] [56]), shown as orange circles.
- **ASICs** (e.g., Google TPUs [126] [74] [127]), shown as pink diamonds.

- **Compute Nodes** (e.g., Nvidia DGX systems [128] [129]), shown as red squares.
- **Supercomputers**, represented by the top-ranked systems on the Top500 list [130] [131], shown as blue squares.
- **Academic research prototypes,** reported in research papers [132] [133] [134] [135] [136], shown as green triangles.

In **Fig. 13**, only AI computing systems capable of scaling up to support LLM workloads are shown, in order to highlight mainstream trends. However, several other types of AI computing platforms are not included, such as *FPGAs* [137] [138], *neuromorphic processors* [139], *photonics computing systems* [140] and *quantum AI platforms* [141].

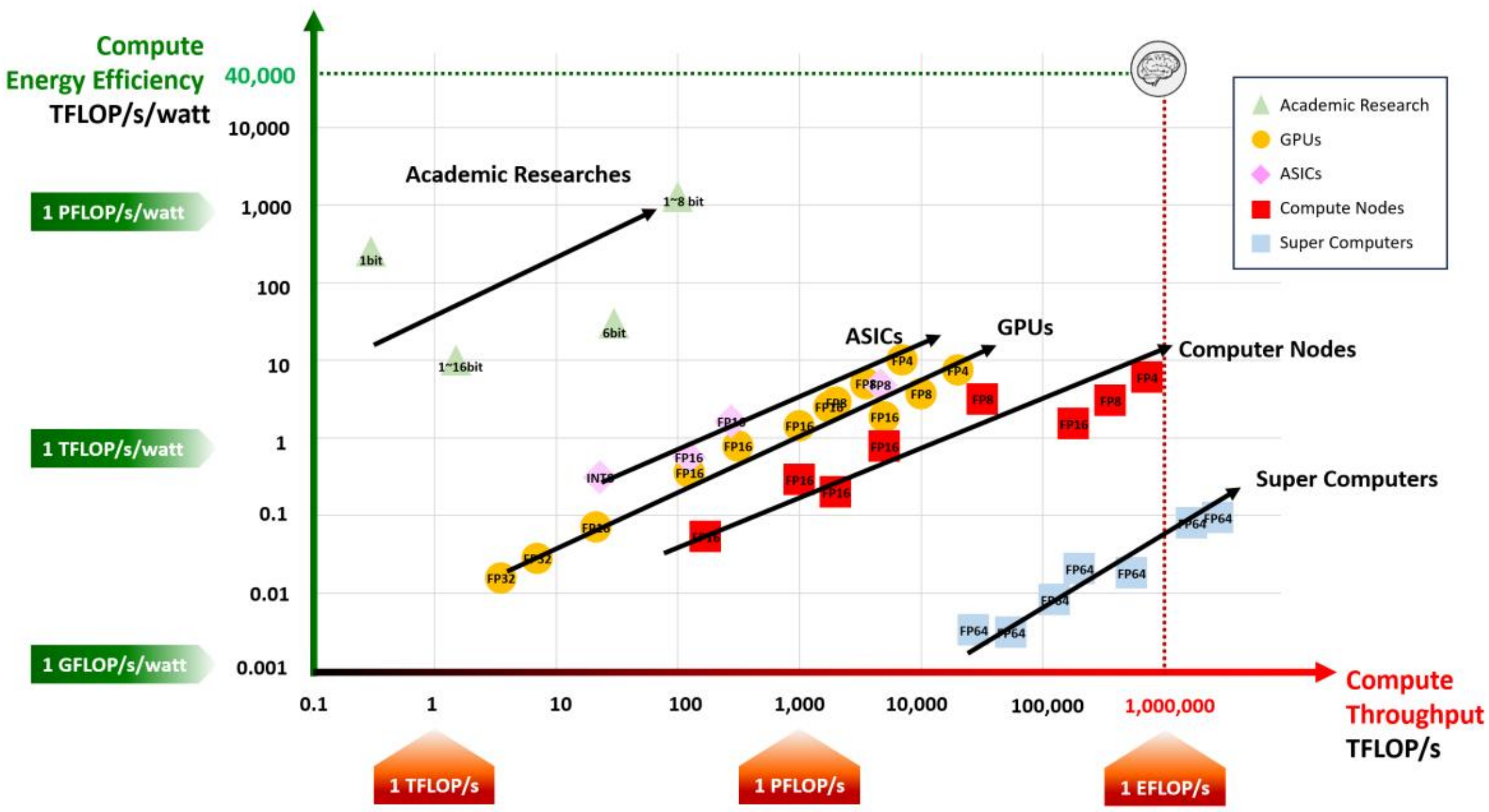

**Fig. 13** Evolution of AI Computing Chips and Systems (by Types)

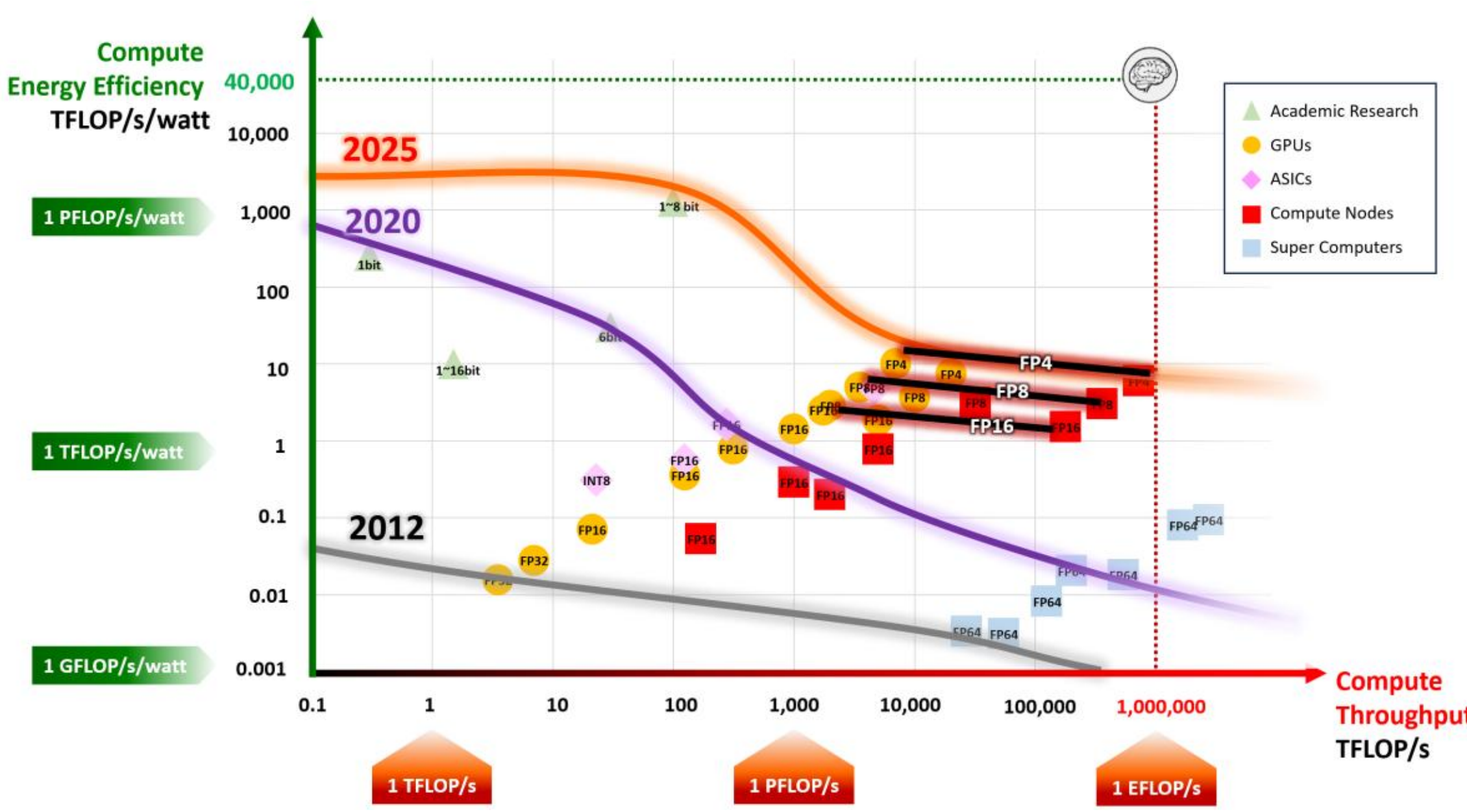

**Fig. 14** Evolution of AI Computing Chips and Systems (by Years)

Numerical precision formats (FP64, FP32, FP16, FP8, FP4, INT8) are labeled within the markers. Since modern GPUs typically support multiple formats (e.g., FP16, FP8, FP4), their performance is plotted separately for each. GPU-based compute nodes are treated in the same way. For supercomputers, FP64 is used because the Top500 rankings are based on the Linpack benchmark. In contrast, academic research results are labeled according to their specific bit formats. For certain specialized approaches, such as photonic computing [65], no numerical precision format is assigned in the figure.

In Fig. **13** the X-axis represents compute throughput (TFLOP/s), and the Y-axis represents compute energy efficiency (TFLOP/s/W), both plotted on logarithmic scales. For reference, the human brain is shown in the upper-right corner. It consumes approximately 25 W of power, derived from an estimated daily energy consumption of 516 kcal [142]:

$$\mathbf{516\ kcal/day} \times 4184\ \text{J/kcal} \div 86400\ \text{s/day} \approx \mathbf{24.99\ W} \quad (1)$$

The brain is further estimated to perform on the order of $10^{18}$ operations per second [143] [144], yielding an energy efficiency of roughly 40,000 TFLOP/s/W.

**Fig. 14** depicts the architectural evolution of AI computing across three milestones—2012, 2020, and 2025:

- **2012 (AlexNet).** The release of AlexNet, trained on GPUs, marked a breakthrough in deep learning. GPUs quickly became the primary engines for AI training, but at this stage were not optimized for neural workloads. The gray trend line highlights the 2012 landscape, where GPU energy efficiency lagged the human brain by six to seven orders of magnitude.

- **2020 (Transformers and LLMs).** The release of the Transformer architecture in 2017, followed by OpenAI's GPT [145] and Google's BERT [146] in 2018, marked the beginning of the era of large language models (LLMs) and drove exponential growth in compute demand. The purple trend line highlights the 2020 landscape, where GPUs and TPUs—optimized through a combination of process scaling and domain-specific architectures (DSAs)—delivered rapid improvements in both performance and energy efficiency.

- **2025 (Current State).** The orange trend line highlights today's landscape. The right-hand region reflects extreme compute throughput enabled by scale-out systems interconnecting hundreds of thousands of chips, though with declining efficiency due to interconnect and data movement overheads. The left-hand region represents modest-scale systems, where energy efficiency improves through specialized chip designs (e.g., compute-in-memory) and ultra-low-bit formats (e.g., 1-bit). Notably, GPUs with different number representation (FP4, FP8 and FP16) yield different energy efficiency.

Overall, scaling compute throughput is relatively straightforward via large-scale scale-out clusters, but improving energy efficiency remains the greater challenge. The human brain sustains advanced intelligence with only ~25 W, whereas today's top-ranked supercomputer consumes ~29.58 MW [147] yet still falls short of brain-level efficiency. This striking disparity underscores the vast headroom for innovation in both neural architectures and AI system design.

In summary, the evolution of the Link Layer underscores a core challenge for AI computing: scaling throughput is achievable, but sustaining energy efficiency is far more difficult. While Scale-Up and Scale-Out drive raw performance, the true bottleneck lies in enabling the Link Layer to deliver scalable, reliable, and energy-efficient interconnects. Overcoming this limitation will be essential for advancing AI systems beyond today's data center scale.

C. Evolution of Layer 3 Neural Network Layer

The Neural Network Layer forms the core of modern AI systems. Over the past decade, various neural network architectures have been developed, including CNNs, RNNs, LSTMs, and GANs. Today, the dominant architecture is the **Transformer model**, which underpins large-scale language models (LLMs). Another widely adopted architecture is the **Diffusion Model**, which powers generative AI applications such as image, video, and content creation.

LLMs have further diversified into multiple sub-categories. Beyond text processing, some LLMs handle multimodal signals such as images, video, audio, and music, and are referred to as **multimodal LLMs (or LMMs, Large Multimodal Models)** [148]. When combined with computer vision capabilities, they are often termed **Vision-Language Models (VLMs)** [149]. More advanced systems that integrate vision, language, and action for behavioral control are referred to as **Vision-Language-Action (VLA)** [150] **models**. For simplicity, this paper collectively refers to these models as LLMs.

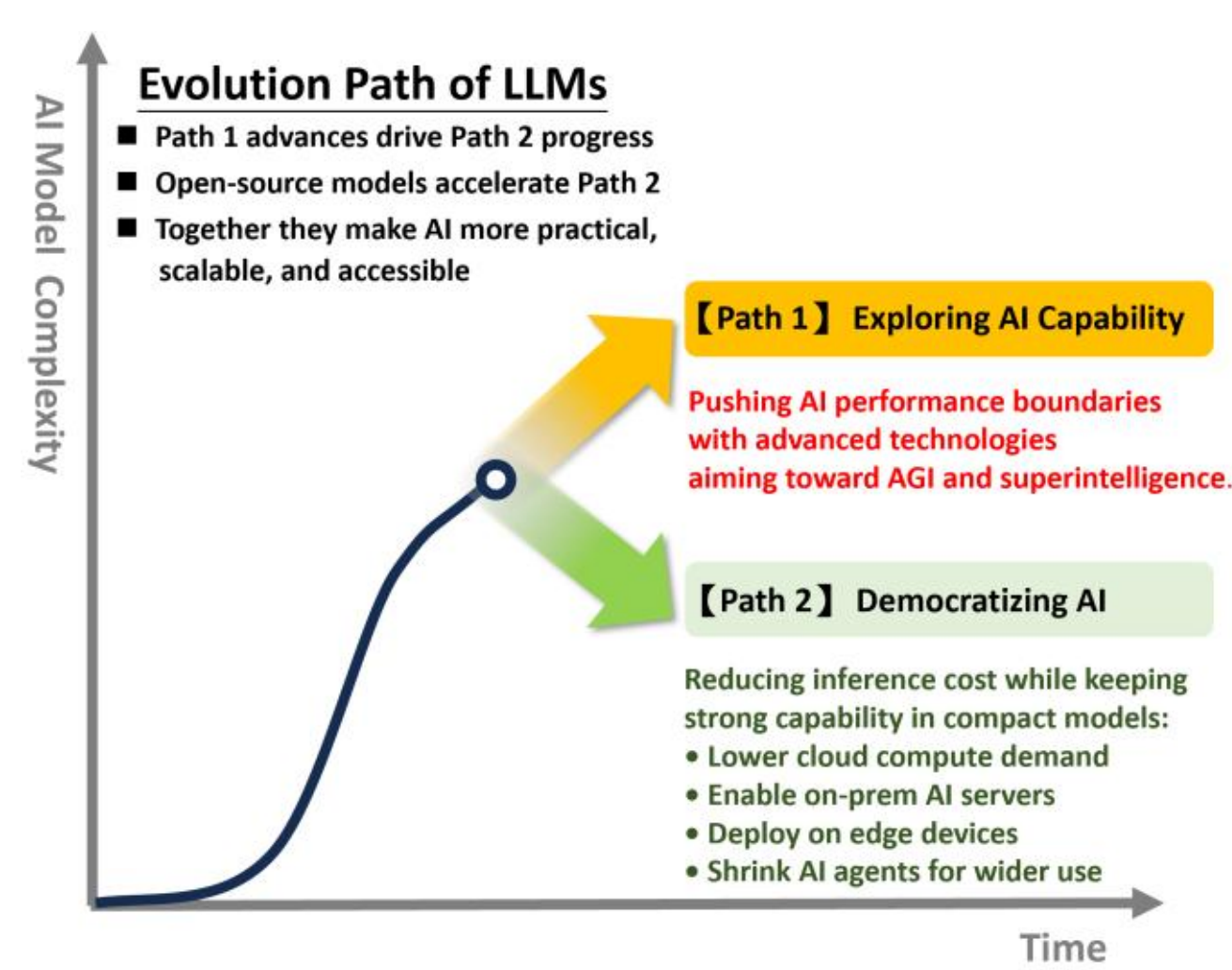

**Fig. 15** Two Paths for LLM Evolution

Building on the earlier discussion of Phases 1 through 3, the future evolution of LLMs is not confined to a single trajectory. At least two complementary development paths can be identified, as shown in **Fig. 15**.

- **Path 1 – Exploring AI Capability:**
  This path pursues ever-greater intelligence, attracting major investments in compute and cutting-edge research. Companies such as OpenAI, Google, Anthropic, and xAI are pushing toward artificial general intelligence (AGI) [151] or even artificial superintelligence (ASI) [152]. However, because training at this scale requires extraordinary computational resources, this path is limited to a small number of well-funded organizations.

- **Path 2 – Democratizing AI:**
  This path emphasizes broad, practical applications. Most real-world use cases do not require full-scale AI functionality. Instead, **knowledge distillation** [18] techniques are used to derive smaller, more efficient models from large teacher LLMs. Although distillation itself consumes additional compute, the resulting lightweight models dramatically reduce inference costs, power consumption, and hardware requirements. This makes AI more accessible to diverse users and industries. The key benefits include:
  - Reduced energy consumption in cloud servers.
  - Deployment on on-premises servers, reducing risks of sending confidential data to the cloud.
  - The ability for sub-billion-parameter models to run on low-cost hardware and edge devices.
  - Lower resource requirements for building AI agents, facilitating the growth of agent-based ecosystems.

The two paths reinforce each other: as Path 1 produces increasingly capable LLMs, Path 2 benefits by distilling those capabilities into smaller, more practical models.

**Fig. 16** illustrates this with an analogy. Path 1 produces full-scale models, just like professors with encyclopedic knowledge across science, humanities, business, and the arts, as well as fluency in multiple human and programming languages. However, just as not every task requires a professor, not every AI application requires a full-scale LLM. Through distillation, **student models** can be created:

- **Medium-sized models** (like graduate students) retain broad expertise in selected domains and are suitable for cloud deployment, balancing capability with efficiency.

- **Small-sized models** (like high school students) focus on narrow subject areas, greatly reducing resource requirements and making them ideal for edge devices.

This dual evolution—toward both cutting-edge research and widespread practical deployment—will define the trajectory of the Neural Network Layer in the coming years.

In summary, the evolution of the Neural Network Layer reflects a dual trajectory. **Path 1** continues to push the frontier of AI capability, driving large-scale investments in training ever-larger LLMs in pursuit of AGI and superintelligence. **Path 2**, in parallel, focuses on democratizing AI through distillation into smaller, more efficient models that enable broad deployment across cloud, enterprise, and edge environments. These two paths are not in conflict but complementary: advances at the frontier feed directly into lightweight models that power practical applications. Together, they illustrate how the Neural Network Layer is evolving from singular breakthroughs into a diverse ecosystem of models—ranging from frontier-scale LLMs to resource-efficient variants—that collectively shape the future trajectory of AI.

### *D. Evolution of Layer 4 Context Layer*

Prompting has become a crucial skill for making LLMs effective. However, as **context memory** is increasingly tasked with longer and more complex instructions—ranging from prompts for test-time compute to functions for AI agents, memory calls, and I/O tokens—the importance of **context engineering** has grown significantly.

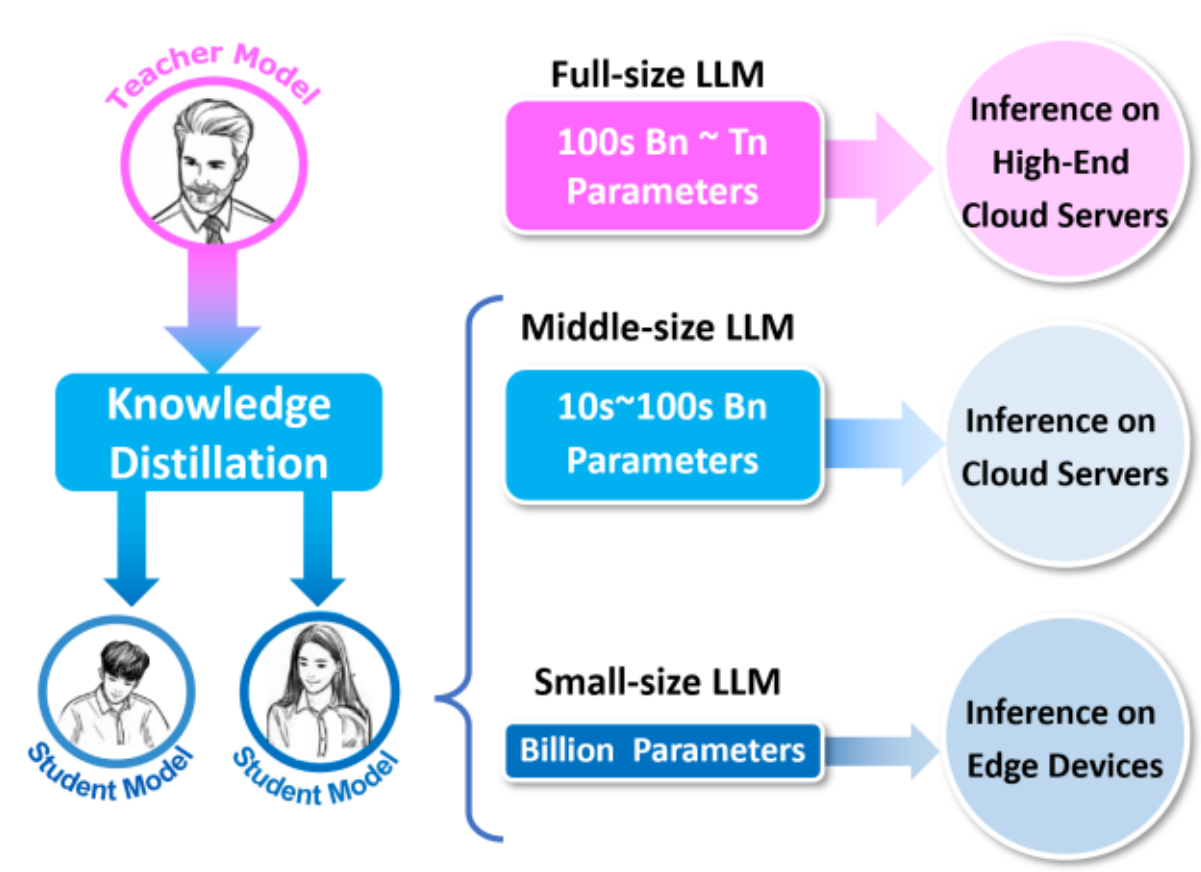

**Fig. 16** Scenarios for LLM Applications: Teacher Model and Student Model

**Fig. 17 (a)** illustrates the typical structure of an LLM. After receiving a prompt and relevant information—converted into tokens and placed into context memory—the LLM processes the content and produces output tokens. Each output token is then appended to the context, helping guide the generation of subsequent tokens.

**Fig. 17 (b)** shows the expanded role of context after applying context engineering. The **prompt section** may include system and user prompts, as well as reasoning-related prompts for test-time compute (e.g., chain-of-thought or tree-of-thought reasoning). The **memory section** incorporates both short-term memory (to store state and interaction history) and long-term memory (to ensure consistency and incorporate documentation or RAG data). Increasingly, LLMs are multimodal and action-enabled, processing signals beyond text (e.g., audio, video) and generating network tokens or action tokens for device and robot control. Additionally, to interact with external tools, context may include protocol-level information.

For comparison, **Fig. 17 (c)** depicts memory in a traditional processor. Several key distinctions highlight the paradigm shift between conventional memory and AI context memory:

- **Tokens:** In AI systems, tokens are the fundamental unit to represent prompts, text and various data, while Traditional processors store program and data in binary form.

- **Attention:** Execution in LLMs is governed by the Transformer's attention mechanism, influenced by both trained parameters and context content. This results in a dynamic, sometimes non-deterministic sequence of operations. Traditional processors, however, execute instructions sequentially as dictated by a program counter, producing deterministic results.

- **Specifications vs. Programming:** Context memory provides high-level specifications, such as blueprints for intended outcomes, rather than step-by-step instructions. Traditional programming, by contrast, encodes operations explicitly in a bottom-up manner.

- **Context Engineering:** Larger context windows allow richer reasoning but also incur extremely high computational cost. Beyond certain limits, performance degradation known as *context rot* [153] or *context collapse* [25] has been observed. Optimizing information layout within finite context memory to maximize attention efficiency has thus become a key research focus. Several techniques include:

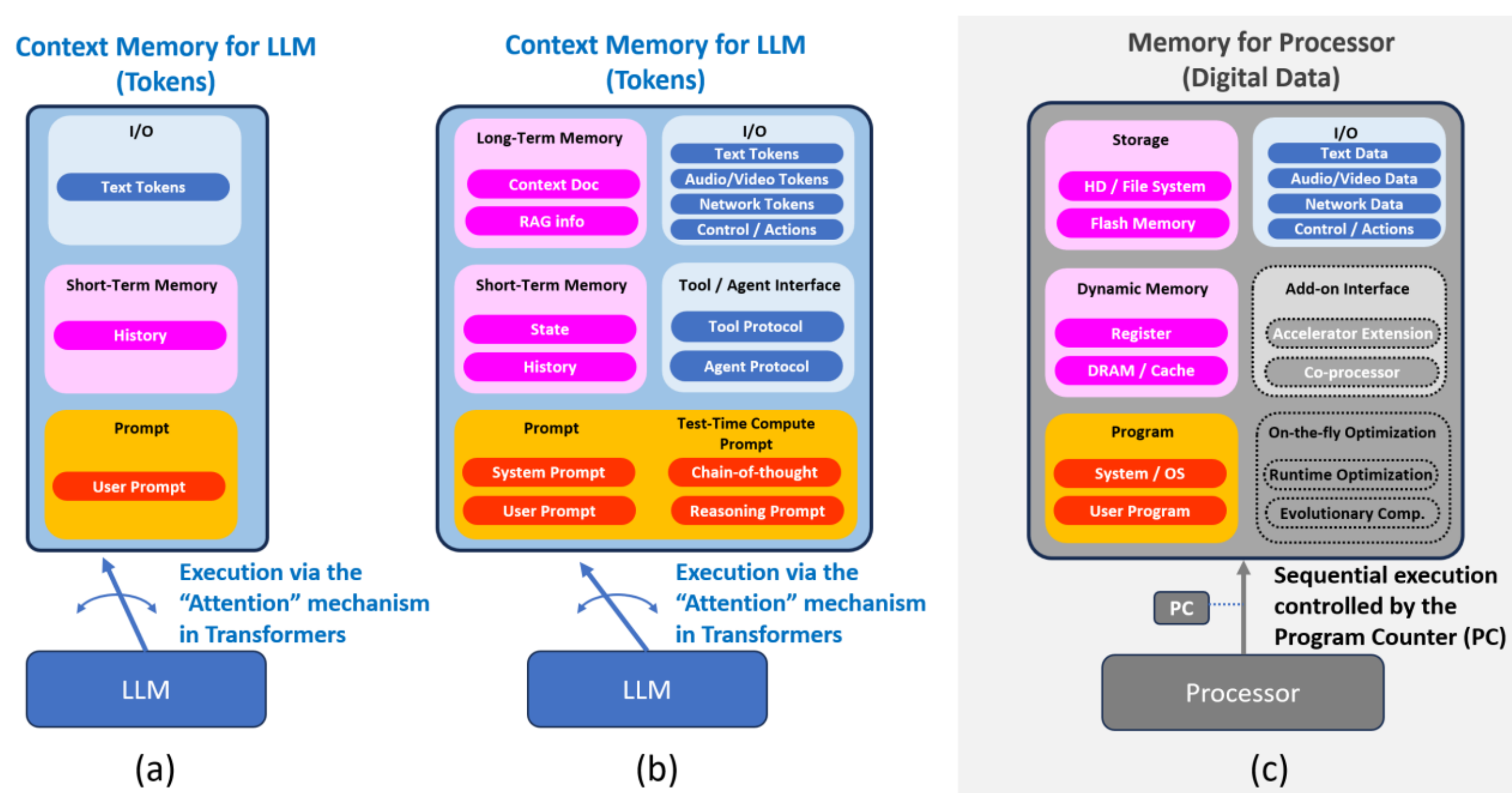

**Fig. 17** Examples for Context Memory Structure for LLMs
(a) Typical Context Memory (b) Context Memory after Context Engineering
(c) Memory Structure in Traditional Processor (for comparison)

(i) **Soft Prompting** – Use compact prompts retaining high-level semantic meaning from natural-language prompts [154].

(ii) **Prompt Compression** – condensing prompts into condensed gist tokens [155].

(iii) **Context Compression** – reducing redundancy in context through embedding compressors [156], in-context compression [157], or compression by vision-token-based optical compression [158].

Context compression differs from conventional memory compression in three ways:

(1) Decompression before execution is unnecessary, since attention can operate directly on compressed representations.

(2) Heterogeneous representation types can coexist within compressed contexts, such as vision-token-based optical compression for text.

(3) Progressive information loss is tolerable—analogous to human memory decay over time [158].

- **On-the-Fly Self-Optimization:** LLMs can dynamically refine their reasoning using self-optimization on context memory, such as dynamic cheat-sheet generation within agentic context frameworks [25]. Traditional computing has limited parallels—e.g., genetic algorithms or just-in-time (JIT) compilation—which affect only isolated modules. In contrast, LLM self-optimization through context memory influences the entire AI system, substantially enhancing its capability.

In summary, the evolution of the **Context Layer** underscores its central role in test-time compute and future agentic AI. Unlike traditional memory, which passively stores code and data, context memory actively shapes reasoning through attention. As prompts, multimodal inputs, and agent interactions accumulate, the demand for larger and more efficient contexts grows. Yet expanding context windows increases computational cost and risks degradation such as context rot. Future progress will depend on **context engineering**—optimizing what to store, how to represent it, and how it guides reasoning. Thus, the Context Layer is no longer a passive storage element but the dynamic workspace where intelligence emerges.

Recent surveys have systematically analyzed the challenges and techniques of context length extension in large language models, highlighting how attention mechanisms, memory organization, and retrieval strategies critically shape the scalability and efficiency of the Context Layer [19].

*E. Evolution of Layer 5, 6, 7*

Currently, the layers above Layer 5—including the **Agent Layer (Layer 5)**, the **Orchestrator Layer (Layer 6)**, and the **Application Layer (Layer 7)**—are still in their early stages of development. Their overarching goal is to extend AI capabilities beyond single-model LLM functions and to achieve higher levels of intelligence through the collaboration of multiple AI agents. Together, these layers will broaden the adoption of AI systems across industries, ultimately shaping an **AI-based Ecosystem**.

**Table 1.** AI Context Memory vs. Traditional Memory

| Aspect | AI System Context Memory | Traditional Processor Memory |
|---|---|---|
| **Granularity** | Tokens (prompts, Text, Images, Audio, *etc.*) | Bits, Bytes (Program, Data) |
| **Access Mechanism** | Probabilistic, attention-driven access and generation | Deterministic, address-based fetch-and-execute |
| **Role** | Active workspace enabling interaction with LLMs for self-optimization | Passive storage under uni-directional control by software programs |
| **Compression** | (1) Decompression not required — attention operates directly on compressed forms.<br>(2) Heterogeneous representations (e.g., vision-token-based optical compression for text) can coexist.<br>(3) Progressive information loss is tolerable, analogous to human memory decay over time. | Decompression required to restore original information type before operation |
| **Behavioral Impact** | Self-optimized context can modify system behavior (“in the loop”) | Memory contents do not alter program logic (“outside the loop”) |
| **Factors Impacting Performance** | Self-optimized context affects LLM performance (output quality) | Memory bandwidth and latency affect throughput (output quantity) |

This new ecosystem will have several important implications:

- **Ecological Regions.** An AI-based ecosystem will consist of numerous AI agents, each specializing in distinct functions. Even small-scale agents can contribute meaningfully, much like species in ecological regions. This lowers the barrier to entry, allowing not only large AI companies but also SMEs, organizations, and individuals to participate in building the ecosystem. While major players may dominate large ecological niches, smaller agents will thrive in more focused domains.

- **Vertical Disintegration.** In this ecosystem, AI agents can interconnect to provide solutions across vertical domains. This enables developers to concentrate on their strengths while integrating into the broader ecosystem. Multiple agents from different providers may deliver similar functions, fostering competition and driving optimization. This mirrors the **vertical disintegration of the semiconductor industry**—where IDMs (integrated device manufacturer) were unbundled into fabless IC design houses, foundries, packaging, and testing providers—resulting in improved efficiency and innovation. A similar approach could create a more dynamic and efficient AI industry supply chain.

- **Value Creation through Specialized AI Agents.** Organizations and individuals with domain expertise will have opportunities to encapsulate their knowledge into AI agents and provide services within the ecosystem. Unlike the traditional model—where high-value data and expertise are absorbed by large AI companies for training—this approach preserves ownership and profit for the creators. It incentivizes professionals to transform their expertise into AI services, enriching the diversity and capacity of the ecosystem.

- **Avoiding Single Points of Failure.** Reliance on a single large-scale LLM creates risks of catastrophic failure if errors, biases, or malicious alterations occur. By contrast, a network of AI agents distributes intelligence and decision-making across decentralized systems. This model, analogous to human societies, allows for committee-style mechanisms, voting, and checks and balances. Faulty agents can be identified and eliminated based on their performance history, creating resilience and reducing systemic risk.

- **New Infrastructure.** If the AI-based ecosystem matures, it may become a foundational infrastructure comparable to the Internet. The scale, however, will be unprecedented.
  - **Users:** Not only humans, but also AI agents, robots, autonomous vehicles, and countless devices will be active participants.
  - **Frequency:** Unlike human-driven infrastructures, AI-driven ecosystems operate at machine speeds, dramatically reducing decision latency. Tasks such as e-commerce transactions could occur at extremely high frequencies, driven by automated agent interactions rather than human deliberation.

Research Directions for Each Layer:

- **Layer 5: Agent Layer**
  - **Protocols and Development Kits:** Several emerging standards, such as Anthropic MCP, Google A2A, OpenAI Swarm, and IBM ACP , are under development to enable interoperability and standardized agent-to-agent communication.
  - **Security and Safety:** As AI agents handle sensitive data and transactions, ensuring robust security frameworks, data privacy, and fault-tolerant operations will be critical.
  - **End-to-End Efficiency and Synergy:** A key challenge lies in maintaining efficiency once multiple agents are integrated. Research is needed to ensure agents can complement each other's strengths, sustain their individual advantages, and achieve true system-level synergy.

- **Layer 6: Orchestrator Layer**
  - **Collaboration:** The orchestrator to coordinate large numbers of agents, assigning tasks and roles to maximize collective performance.
  - **Evaluation and Selection:** Systematic methods are required to evaluate agents, measure reliability, and assign trust scores, similar to credit ratings in human society.
  - **Identity and Anti-Counterfeiting:** Reliable identity verification and anti-spoofing mechanisms will be essential to prevent malicious or counterfeit agents from infiltrating the ecosystem.
  - **Resource Management:** Effective allocation of compute, memory, and communication bandwidth across agents will be necessary, paralleling human resource management in organizations.

- **Lifecycle Management:** Agents will need structured lifecycle management, including versioning, updates, patching, and controlled retirement to ensure reliability and safety

- **Layer 7: Application Layer**
  - **Infrastructure Support:** Large-scale AI applications will depend on robust infrastructures for information, financial, and logistics flows.
  - **User Authorization:** Mechanisms must allow users to authenticate and selectively authorize agents to act on their behalf.
  - **Safety and Emergency Controls:** Oversight mechanisms must monitor agent behavior, prevent unethical or harmful actions, and allow for rapid emergency shutdown when required.
  - **Seamless Service:** As AI inference becomes critical infrastructure, maintaining continuous availability and resilience against disruptions will be a key design objective.

In summary, Layers 5–7 extend AI beyond single models toward ecosystem-level intelligence. Layer 5 (Agent) augments LLMs with memory, planning, tool use, and interaction capabilities. Layer 6 (Orchestrator) coordinates and manages networks of agents, while Layer 7 (Application) embeds AI into real-world systems with safety, reliability, and seamless service. Together, these layers represent a shift from model-centric AI to an ecosystem-centric paradigm that scales across industries and society.

## V. SUMMARY

This article analyzed AI compute architecture through a seven-layer model, tracing its evolution and future trajectory. The key observations are summarized as follows:

Layers 1–2 (Physical & Link): AI training compute has increased by a factor of 100 million over the past decade. While Scale-Up advances improve performance on single compute node, Scale-Out dominates system scaling. Yet, large-scale interconnects introduce significant energy inefficiencies. Inference compute will likely exceed training compute in scale, as future users include not only humans but also AI agents and robots.

Layer 3 (Neural Networks): Two evolutionary paths are evident: (i) scaling LLMs toward ever-higher capability to AGI, and (ii) distilling large models into compact ones for practical deployment. The latter is particularly critical for enabling agents, edge AI, and democratized applications. Beyond single models, the next frontier includes Agentic AI and Physical AI.

Layer 4 (Context): Context memory has become central to reasoning. As it grows in scope—covering prompts, tokens, tools, and multimodal inputs—context engineering is now critical for maximizing performance within finite memory and compute budgets.

Layer 5 (Agents): LLMs augmented with memory, planning, and tool use evolve into agents. The challenge lies in achieving end-to-end efficiency and synergy across interconnected agents, while avoiding single points of failure through decentralized decision-making.

Layer 6 (Orchestrators): Orchestrators govern multi-agent systems, akin to resource and lifecycle management in organizations. Systematic evaluation, identity verification, and trust assignment will be essential for building resilient ecosystems.

Layer 7 (Applications): The emerging AI-based ecosystem extends beyond models and agents to encompass humans, devices, robots, and infrastructure—including information, financial, and logistics flows. Its synergy across humans, AI, and machines could drive a historic leap in productivity. Reliability, safety, and uninterrupted inference services will be vital as AI matures into essential infrastructure.

In conclusion, AI exhibits unprecedented potential but also demands extraordinary resources. Its future impact may surpass prior industrial revolutions, reshaping productivity, economics, and society. The central challenge is not only technical scaling, but also establishing sustainable economic and governance systems to support growth. If achieved, AI will evolve from powerful models into a resilient, global ecosystem—an enduring foundation for the next era of human progress.

## ACKNOWLEDGMENT

The author would like to thank colleagues at MediaTek, the MediaTek Advanced Research Center (MARC), MediaTek Research, National Taiwan University, National Yang Ming Chiao Tung University, National Tsing Hua University, IEEE CASS, SSCS, Taipei section and Tainan section for their support and collaboration. The author would like to thank reviewers for their valuable comments and suggestions. The clip art and illustration used in figures were generated with OpenAI ChatGPT and DALL·E. The English presentation of this paper was refined with the assistance of ChatGPT.

# APPENDIX

**Table 2** Performance and Energy Efficiency for AI Computing

| Year | Type | Name | Number Format | Performance (TFLOP/s) | Energy Efficiency (TFLOP/s/W) |
|---|---|---|---|---|---|
| | *Brain* | *Human Brain* | | *1,000,000.000* | *40,000.000* |
| 2012 | GPU | K20x | FP32 | 3.524 | 0.016 |
| 2012 | Super Computer | TITAN | FP64 | 27,112.550 | 0.003 |
| 2015 | GPU | M40 | FP32 | 7.000 | 0.028 |
| 2015 | ASIC | TPUv1 | INT8 | 23.000 | 0.310 |
| 2015 | Super Computer | Tianhe-2 | FP64 | 54,902.400 | 0.003 |
| 2016 | GPU | P100 | FP16 | 21.200 | 0.071 |
| 2016 | Compute Node | DGX-1 | FP16 | 170.000 | 0.053 |
| 2016 | Super Computer | Tianhe-2 | FP64 | 125,435.900 | 0.008 |
| 2017 | GPU | V100 | FP16 | 125.000 | 0.357 |
| 2017 | Compute Node | DGX-1 | FP16 | 1,000.000 | 0.286 |
| 2017 | Research | Envision | 1~16bit | 1.500 | 10.000 |
| 2018 | Research | BinarEye | 1bit | 0.300 | 250.000 |
| 2018 | ASIC | TPUv3 | FP16 | 123.000 | 0.560 |
| 2018 | Compute Node | DGX-2 | FP16 | 2,000.000 | 0.200 |
| 2018 | Super Computer | Summit | FP64 | 200,794.880 | 0.021 |
| 2019 | Research | IMC CNN | 6bit | 30.000 | 30.000 |
| 2020 | GPU | A100 | FP16 | 312.000 | 0.780 |
| 2020 | Compute Node | DGX- A100 | FP16 | 5,000.000 | 0.769 |
| 2020 | Super Computer | Fugaku | FP64 | 537,210.000 | 0.018 |
| 2021 | ASIC | TPUv4 | FP16 | 275.000 | 1.620 |
| 2022 | GPU | H100 | FP8 | 2,000.000 | 2.857 |
| 2022 | Compute Node | DGX-H100 | FP8 | 32,000.000 | 3.137 |
| 2022 | GPU | H100 | FP16 | 1,000.000 | 1.429 |
| 2022 | Super Computer | Frontier | FP64 | 1,685,650.000 | 0.080 |
| 2022 | Research | ReRAM CIM | 1~8 bit | 100.000 | 1,286.400 |
| 2023 | GPU | B100 | FP16 | 1,750.000 | 2.500 |
| 2023 | GPU | B100 | FP8 | 3,500.000 | 5.000 |
| 2023 | GPU | B100 | FP4 | 7,000.000 | 10.000 |
| 2024 | GPU | GB200 | FP16 | 5,000.000 | 1.852 |
| 2024 | GPU | GB200 | FP8 | 10,000.000 | 3.704 |
| 2024 | GPU | GB200 | FP4 | 20,000.000 | 7.407 |
| 2024 | Compute Node | DGX-GB200 NVL72 | FP16 | 180,000.000 | 1.500 |
| 2024 | Compute Node | DGX-GB200 NVL72 | FP8 | 360,000.000 | 3.000 |
| 2024 | Compute Node | DGX-GB200 NVL72 | FP4 | 720,000.000 | 6.000 |
| 2024 | Super Computer | El Capitan | FP64 | 2,746,380.000 | 0.093 |
| 2025 | ASIC | TPUv7 | FP8 | 4,614.000 | 4.800 |

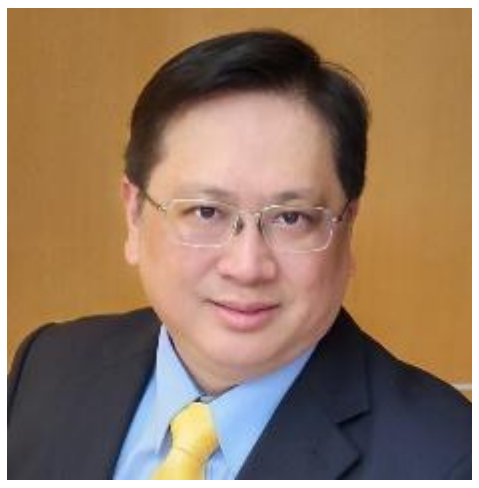

**Bor-Sung Liang** *(Senior Member, IEEE)* is a Senior Director, Corporate Strategy & Strategic Technology of MediaTek Inc., Hsinchu Science Park, Taiwan, and a Director of the Board, MediaTek Foundation. He is also concurrently serving as a Visiting Professor at CSIE (Department of Computer Science and Information Engineering) and GIEE (Graduate Institute of Electronics Engineering), EECS (College of Electrical Engineering and Computer Science) and GSAT (Graduate School of Advanced Technology) in National Taiwan University, as well as a Visiting Professor at ECE (College of Electrical and Computer Engineering) in National Yang Ming Chiao Tung University. Dr. Liang is a Director of IEEE Taipei Section, and an IEEE CASS (Circuits and Systems Society) Industrial Distinguished Lecturer (2025-2026). He was the Chair of IEEE CASS Taipei Chapter (2023-2024). He is also the executive director of Taiwan IC Industry & Academia Research Alliance (TIARA).

He received his Ph.D degree from Institute of Electronics, National Chiao Tung University, and graduated from EMBA, College of Management, National Taiwan University. Dr. Liang has received several important awards, such as Ten Outstanding Young Persons, Taiwan, R.O.C., National Invention and Creation Award on Invention (for three times, one Gold Medal and two Silver Medals) from Intellectual Property Bureau of the Ministry of Economic Affairs, Taiwan, Outstanding Youth Innovation Award of Industrial Technology Development Award from Department of Industrial Technology of the Ministry of Economic Affairs, Taiwan, Outstanding ICT Elite Award of ICT Month, R.O.C., and K. T. Li Young Researcher Award from Institute of Information & Computing Machinery and ACM Taipei/Taiwan Chapter. His major research fields are AI computing architecture, digital IC design, processor architecture, quantum computing, and technology strategy.